\documentclass[journal,twoside]{IEEEtran}

\usepackage{algorithmicx}
\usepackage{algorithm}% http://ctan.org/pkg/algorithms
\usepackage{algpseudocode}% http://ctan.org/pkg/algorithmicx
\usepackage{times}
\usepackage{footnote}

\usepackage{multicol}
\newenvironment{experiment}
  {\begin{enumerate}}
    {\end{enumerate}}
  
\newenvironment{simulation}
  {\begin{enumerate}}
  {\end{enumerate}}

\newcommand{\ssRegion}{{\set{\text{ssr}}}}

\newcommand{\cRegion}{\set{cr}}

\newcommand{\coefPF}{a} % The scalar to predict the peak contact force from impulse and impact duration.

\newcommand{\highlight}[1]{%
  \colorbox{gray!20}{$\displaystyle#1$}}

\newcommand{\highlightblue}[1]{%
  \colorbox{blue!20}{$\displaystyle#1$}}

\newcommand{\ajDim}{n}
\newcommand{\fbDim}{6}
\newcommand{\jsDim}{(\ajDim+\fbDim)}
\newcommand{\actuationMatrix}{\text{B}}
\usepackage{xspace}
\newcommand{\unitPos}{~m\xspace}

\newcommand{\unitVelTS}{~m/s\xspace}

\newcommand{\unitAccTS}{~m/s$^2$\xspace}

\newcommand{\unitForce}{~N\xspace}

\newcommand{\unitMS}{~ms\xspace}

\usepackage{multirow}

\usepackage[utf8]{inputenc}
\usepackage[english]{babel}
\usepackage[dvipsnames]{xcolor}

\usepackage{listings}
\usepackage{color}
\usepackage{tikz}
\usepackage{tikz-layers}
\usepackage{array}
\newcolumntype{C}[1]{>{\centering\arraybackslash}p{#1}}
\usetikzlibrary{shapes.geometric,automata,positioning,arrows,shadows,,backgrounds,calc}

\definecolor{frameColor}{RGB}{128,128,128}
\definecolor{initialStateColor}{RGB}{10,100,10}
\definecolor{impactStateColor}{RGB}{0,255,0}
\definecolor{admittanceStateColor}{RGB}{0,0,255}

\definecolor{resetStateColor}{RGB}{10,100,10}

\definecolor{stateColor}{RGB}{51,204,204}

\definecolor{controllerColor}{RGB}{10,10,255}
\definecolor{plannerColor}{RGB}{10,150,150}
\definecolor{estimatorColor}{RGB}{10,200,10}
\definecolor{robotColor}{RGB}{255,0,0}
\definecolor{modelColor}{RGB}{180,100,10}

\usepackage{kantlipsum}

\usepackage{setspace} 

\usepackage{mathtools}
\usepackage{amssymb}
\usepackage{graphicx}
\usepackage[font=normal,labelfont=bf]{caption}
\usepackage{url}
\usepackage{accents} %for symbols below a character, e.g. \underaccent{\bar}
\usepackage[colorlinks=true,hyperindex=true,bookmarksnumbered,bookmarks=true]{hyperref} % Turn on "bookmarks" and "hyperindex" to generate outlines. 

\newcommand{\setDef}[2]{
  \defeq \{#1 : #2 \}
}

\newcommand{\quickEq}[2]{
  \begin{equation}
    \label{#1}
    {#2}
  \end{equation}
}

\newcommand{\unitHz}{~Hz}

\newcommand{\setSymbol}{\mathcal{X}}
\newcommand{\set}[1]{\setSymbol_{#1}}

\usepackage{pifont}% http://ctan.org/pkg/pifont
\newcommand{\contactVel}{\bs{v}}

\newcommand{\iim}{W}

\newcommand{\optimal}[1]{{#1}^{*}}

\newcommand{\postImpact}[1]{{#1}^{+}}
\newcommand{\preImpact}[1]{#1^{-}}
\newcommand{\coefR}{c_{\text{r}}}
\newcommand{\coefF}{\mu}

\newcommand{\nz}{\text{\bf n}}

\newcommand{\fConeNum}{N_\coefF}

\newcommand{\basisVec}[1]{\widehat{\bs{\bs{#1}}}}

\definecolor{csrStateColor}{RGB}{255,255,153}
\definecolor{scrStateColor}{RGB}{153,204,0}
\definecolor{crStateColor}{RGB}{153,204,155}

\newcommand{\jointPosition}{\bs{\theta}}

\newcommand{\inertialFrame}{\text{O}}
\newcommand{\inertialWrench}{{\wrench_{\inertialFrame}}}
\newcommand{\inertialForce}{{\force_{\inertialFrame}}}
\newcommand{\inertialTorque}{{\torque_{\inertialFrame}}}
\newcommand{\comFrame}{\com}

\newcommand{\gravityandcoriolis}{\mathbf{N}}
\newcommand{\gForce}{\bs{g}}
\newcommand{\gValue}{\text{9.81}}

\newcommand{\twoNorm}[1]{
  {{\left\lVert#1\right\rVert}^2}
}

  \newcommand{\cross}[2]{ #1 \times #2 }

\newcommand{\state}{\bs{x}}

\newcommand{\innerP}[2]{
  \transpose{#1}#2
}

\newcommand{\wrenchAS}[2]{S}

\newcommand{\inertiaMatrix}{\text{M}}

\newcommand{\agg}[3]{
\sum^{#2}_{#1=1}{#3}
}

\newcommand{\vc}[1]{\bs{#1}}

\newcommand{\vectorTwo}[2]{
  \begin{bmatrix}
    #1\\
    #2
\end{bmatrix}
}

\newcommand{\vectorTwoRow}[2]{
  [#1^\top, #2^\top]^\top
}

\newcommand{\matrixTwo}[4]{
  \begin{bmatrix}
    #1 & #2\\
    #3 & #4
\end{bmatrix}
}

\newcommand{\rotation}[2]{R_{#1 #2}}

\newcommand{\rotationInv}[2]{R^{\top}_{#1 #2}}

\newcommand{\translationSkew}[2]{\skewMatrix{\vc{p}}_{#1 #2}}

\usepackage{dsfont}
\newcommand{\identityMatrix}{\mathds{1}} %matrix % alternativ \mathds{1} or {\mathbb{1}}
\newcommand{\idMatrix}[1]{{\identityMatrix_{#1}}} %matrix % alternativ \mathds{1} or {\mathbb{1}}

\newcommand{\zeroVector}{\mathbf{0}} %matrix % alternativ \mathds{0} or {\mathbb{0}} or \mathcal{O}
\newcommand{\zeroMatrix}{0} %matrix % alternativ \mathds{0} or {\mathbb{0}} or \mathcal{O}

\newcommand{\twistTransform}[2]{
  \textcolor{geometricColorVariable}{
    \adgInv{#1}{#2}
    }
  }

\newcommand{\twistTransformDef}[2]{
  \adgInvDef{#1}{#2}
}

\newcommand{\adgInv}[2]{
  \textcolor{geometricColorVariable}{
  Ad^{-1}_{g_{#1#2}}
}}

\newcommand{\adgInvDef}[2]{
  \textcolor{geometricColorVariable}{
  \begin{bmatrix}
    \rotationInv{#1}{#2} & -\rotationInv{#1}{#2}\translationSkew{#1}{#2} \\
    \zeroMatrix & \rotationInv{#1}{#2}
\end{bmatrix}
}}

\newcommand{\wrench}{\bs{W}}
\newcommand{\cwc}{\text{C}}
\newcommand{\cwcFrame}[2]{{^{#1}\cwc_{#2}}}

\newcommand{\wrenchFrame}[2]{{^{#1}\wrench_{#2}}} % The arguments are: (1)Frame (2) Application point
\newcommand{\cone}[1]{{\text{K}_{#1}}}
\newcommand{\generator}[1]{{\bs{\lambda}_{#1}}}
\newcommand{\generatorScalar}[1]{{\lambda_{#1}}}

\newcommand{\wrenchC}[2]{
\begin{bmatrix}
#1\\
#2
\end{bmatrix}
}

\newcommand{\wrenchR}[2]{
[#1^\top, #2^\top]^\top
}

\newcommand{\reff}[1]{#1^{\text{ref}}}

\newcommand{\error}[1]{\bs{e}_{#1}}
\newcommand{\errord}[1]{\dot{\bs{e}}_{#1}}
\newcommand{\errordd}[1]{\ddot{\bs{e}}_{#1}}

\newcommand{\bs}{\boldsymbol}
\newcommand{\backfill}{\hfill\(\blacksquare\)}

\newcommand{\skewMatrix}[1]{\widehat{#1}}

\newcommand{\vectripleexpan}[3]{
(\inner{#1}{#3}){#2} - (\innerP{#1}{#2}){#3}
}

\newcommand{\inner}[2]{#1\cdot#2} % Inner product 
\newcommand{\transpose}[1]{{#1}^\top}

\newcommand{\pseudoInverseRowDef}[1]{\transpose{#1}\inverse{( #1 \transpose{#1} )}}

\newcommand{\inverse}[1]{{#1}^{-1}}

\newcommand{\resultantTorque}[4]{#2 + (#3 - #4)\times #1}
\newcommand{\resultantTorqueTwo}[4]{#2 + \pointer{#4}{#3}\times #1}

\newcommand{\cframe}[1]{\mathcal{F}_{#1}}

\newcommand{\mass}{\text{m}}

\newcommand{\numContacts}{{\text{m}_{\text{sc}}}}

\newcommand{\contactPoint}{\bs{p}}

\newcommand{\point}[1]{\boldsymbol{q}_{#1}}

\newcommand{\force}{\bs{f}}

\newcommand{\comPlane}{\com_{x,y}}
\newcommand{\pendulumC}{\text{w}}
\newcommand{\pendulumCDef}{\sqrt{\frac{g}{\com_z}}}
\newcommand{\comVel}{\dot{\com}}
\newcommand{\comVelPlane}{\dot{\com}_{x,y}}

\newcommand{\torque}{\bs{\tau}}

\newcommand{\pGain}{k_p}
\newcommand{\dGain}{k_d}

\newcommand{\pole}[1]{p_{#1}}

\newcommand{\angularMomentum}{\mathcal{L}}
\newcommand{\angularMomentumD}{\dot{\angularMomentum}}
\newcommand{\linearMomentum}{P}
\newcommand{\linearMomentumD}{\dot{\linearMomentum}}

\newcommand{\impactDuration}{\delta t}

\newcommand{\jump}{\Delta}

\newcommand{\zmp}{{\text{\bf z}}}

\newcommand{\height}[1]{{d_{#1}}}

\newcommand{\dcm}{{\bs{\xi}}}

\newcommand{\dcmxDef}{{\com_x + \frac{\comVel_x}{\pendulumC}}}

\newcommand{\com}{{\text{\bf c}}}
\newcommand{\comd}{{\dot{\com}}}
\newcommand{\comdxy}{\comd_{xy}}
\newcommand{\comdPlane}{{\dot{\com}_{x,y}}}

\newcommand{\picomdxy}{\postImpact{\comd}_{xy}}
\newcommand{\picomdSet}{\set{\postImpact{\comd}_{xy}}}
\newcommand{\picomdVertex}[1]{\set{\postImpact{\comd}_{xy},#1}}
\newcommand{\comdd}{{\ddot{\com}}}

\newcommand{\pointer}[2]{\overrightarrow{#1#2}}
\newcommand{\pointerDef}[2]{\bs{#2}  - \bs{#1}}

\newcommand{\impulse}{\bs{\iota}} %TODO similar to identityMatrix
\newcommand{\supportPolygon}{\set{\mathcal{S}}}
\newcommand{\zmpArea}{{\set{[\zmp]}}}

\newcommand{\comdArea}{\set{[\comd_{xy}]} }

\newcommand{\jangles}{\mathbf{q}}
\newcommand{\jvelocities}{\mathbf{\dot{q}}}
\newcommand{\jaccelerations}{\mathbf{\ddot{q}}}
\newcommand{\jtorques}{\boldsymbol{\tau}}

\newcommand{\jacobian}{J}

\newcommand{\lowerBound}[1]{\underline{#1}}
\newcommand{\upperBound}[1]{\overline{#1}}

\newtheorem{remark}{Remark}[section]

\newtheorem{problem}{Problem}

\usepackage{mathtools}
\newcommand{\defeq}{\vcentcolon=}

\newcommand{\RRv}[1]{\mathbb{R}^{#1}}
\newcommand{\RRm}[2]{\mathbb{R}^{#1 \times #2}}

\definecolor{dkgreen}{rgb}{0,0.6,0}
\definecolor{gray}{rgb}{0.5,0.5,0.5}
\definecolor{mauve}{rgb}{0.58,0,0.82}

\NewDocumentCommand{\cpp}{v}{%
\texttt{\textcolor{blue}{#1}}%
}

\NewDocumentCommand{\probRef}{v}{%
Problem~\ref{#1}}

\NewDocumentCommand{\secRef}{v}{%
Sec.~\ref{#1}}

\NewDocumentCommand{\remarkRef}{v}{%
Remark~\ref{#1}}

\NewDocumentCommand{\tableRef}{v}{%
Table~\ref{#1}}

\NewDocumentCommand{\algRef}{v}{%
Algorithm~\ref{#1}}

\NewDocumentCommand{\lemmaRef}{v}{%
Lemma~\ref{#1}}

\NewDocumentCommand{\theoremRef}{v}{%
Theorem~\ref{#1}}

\NewDocumentCommand{\figRef}{v}{%
  Fig.~\ref{#1}}

\NewDocumentCommand{\appRef}{v}{%
  Appendix~\ref{#1}}

\NewDocumentCommand{\exRef}{v}{%
Experiment.~\ref{#1}}

\NewDocumentCommand{\simRef}{v}{%
Simulation.~\ref{#1}}

\NewDocumentCommand\orderedTwoS{mm}%
  {$<$#1,#2$>$}
\NewDocumentCommand\orderedThreeS{mmm}%
  {$<$#1,#2,#3$>$}

\definecolor{svaColor}{RGB}{7, 160, 2}
\definecolor{uncertainColor}{RGB}{255, 0, 0}

\definecolor{spatialVelColor}{RGB}{189, 38, 96}
\definecolor{geometricColor}{RGB}{66, 126, 245}
\definecolor{bodyVelColor}{RGB}{13, 191, 191}

\definecolor{velColor}{RGB}{50, 205, 50}

\colorlet{svaColorVariable}{svaColor}
\colorlet{geometricColorVariable}{geometricColor}

\newif\ifBlockComment
\BlockCommentfalse %\BlockCommenttrue
\usepackage{subfigure}
\newcolumntype{C}[1]{>{\centering\arraybackslash}p{#1}}
\usetikzlibrary{automata, positioning, arrows, shadows, backgrounds}

\tikzset{
  basic box/.style = {
    shape = rectangle,
    align = center,
    draw  = #1,
    fill  = #1!25,
    rounded corners},
  header node/.style = {
    font          = \strut\footnotesize\ttfamily,
    text depth    = +0pt,
    fill          = white,
    draw},
  header/.style = {%
    inner ysep = +1.5em,
    append after command = {
      \pgfextra{\let\TikZlastnode\tikzlastnode}
      node [header node] (header-\TikZlastnode) at (\TikZlastnode.north) {#1}
      node [span = (\TikZlastnode)(header-\TikZlastnode)]
        at (fit bounding box) (h-\TikZlastnode) {}
    }
  },
  hv/.style = {to path = {-|(\tikztotarget)\tikztonodes}},
  vh/.style = {to path = {|-(\tikztotarget)\tikztonodes}},
  fat blue line/.style = {ultra thick, blue}
}

\begin{document}

% \title{Zero-Step Capture Region for Non-Coplanar Contacts}
\title{Impact-Aware Multi-Contact Balance Criteria}
% You will get a Paper-ID when submitting a pdf file to the conference system
% \author{Author Names Omitted for Anonymous Review. Paper-ID [XXX]}

\author{Yuquan Wang$^{1,2}$, Arnaud Tanguy$^{2}$, and Abderrahmane Kheddar$^{2,3}$,~\IEEEmembership{Fellow,~IEEE}%
\thanks{Manuscript received April 30, 2023; revised Xxxxx XX, 2023; accepted Xxxxxx XX, 20XX. Date of publication Xxxxxxxx X, 20XX; date of current version Xxxxxxxxx X, 20XX. This paper was recommended for publication by Associate Editor X. Xxxxxxxx and Editor Paolo Robuffo Giordano upon evaluation of the reviewers comments.}
\thanks{This work is supported in part by the Research Project I.AM. through the European Union H2020 program (GA 871899).}
\thanks{$^{1}$ Y. Wang is with the Department of Advanced Computing Sciences, Maastricht University, Maastricht, The Netherlands. 
}%
\thanks{$^{2}$ A. Tanguy and A. Kheddar are with the CNRS-University of Montpellier LIRMM, Montpellier, France.
  Y. Wang was in the same lab when this paper was initially drafted.
}%
\thanks{$^{3}$ A. Kheddar is also with the CNRS-AIST Joint Robotics Laboratory, IRL, Tsukuba, Japan.}%
\thanks{Corresponding author: Y. Wang {yuquan.wang@maastrichtuniversity.nl}}%
\thanks{This paper has supplementary video downloadable material available at http://ieeexplore.ieee.org.}
\thanks{Color versions of one or more of the figures in this paper are available online at http://ieeexplore.ieee.org.}
\thanks{Digital Object Identifier 00.0000/XXX.202X.0000000}
}

%\author{\authorblockN{Michael Shell}
%\authorblockA{School of Electrical and\\Computer Engineering\\
%Georgia Institute of Technology\\
%Atlanta, Georgia 30332--0250\\
%Email: mshell@ece.gatech.edu}
%\and
%\authorblockA{Twentieth Century Fox\\
%\and
%\authorblockN{James Kirk\\ and Montgomery Scott}
%\authorblockA{Starfleet Academy\\
%San Francisco, California 96678-2391\\

% avoiding spaces at the end of the author lines is not a problem with
% conference papers because we don't use \thanks or \IEEEmembership

%Homer Simpson\authorrefmark{2},
%Montgomery Scott\authorrefmark{3} and
%Eldon Tyrell\authorrefmark{4}}
%Telephone: (800) 555--1212, Fax: (888) 555--1212}
%\authorblockA{\authorrefmark{4}Tyrell Inc., 123 Replicant Street, Los Angeles, California 90210--4321}}

\markboth{IEEE Transactions on Robotics. Preprint version}%
{Wang \MakeLowercase{\textit{et al.}}: Impact-Aware Multi-Contact Balance Criteria}

\IEEEpubid{0000--0000/00\$00.00~\copyright~202X IEEE}

\maketitle

\begin{abstract}

  % Intentionally applying impacts is challenging for legged robots. By observing the humanoid robot HRP-4 intentionally hitting a wall, we noticed that violating traditional ZMP-based balance criteria will not always lead to a fall.
  % % .  which revealed that violating traditional balance criteria did not always result in a fall.
  % To investigate this phenomenon, we compute the zero-step capture region for non-coplanar contacts, defined as the center of mass (CoM) velocity area, and validate it with push-recovery experiments on the HRP-4 balancing on two feet contacts sustained on a 30-degree ramp and the ground, respectively. To enable on-purpose impacts accounting for three-dimensional friction,
  % we compute the set of candidate post-impact CoM velocities and restrict the entire set within the CoM velocity area to ensure balance is maintained with the sustained contacts. We illustrate the maximum contact velocity for various HRP-4 stances in simulation, indicating potential for integration into other task-space whole-body controllers or planners. This study is the first to address the challenging problem of applying an intentional impact with a kinematic-controlled humanoid robot on non-coplanar contacts.

  Intentionally applying impacts while maintaining balance is challenging for legged robots. This study originated from observing experimental data of the humanoid robot HRP-4 intentionally hitting a wall with its right arm while standing on two feet. Strangely, violating the usual zero moment point balance criteria did not systematically result in a fall. To investigate this phenomenon, we propose the zero-step capture region for non-coplanar contacts, defined as the center of mass (CoM) velocity area, and validated it with push-recovery experiments employing the HRP-4 balancing on two non-coplanar contacts.
  To further enable on-purpose impacts, we compute the set of candidate post-impact CoM velocities accounting for frictional-impact dynamics in three dimensions, and restrict the entire set within the CoM velocity area to maintain balance with the sustained contacts during and after impacts. We illustrate the maximum contact velocity for various HRP-4 stances in simulation, indicating potential for integration into other task-space whole-body controllers or planners. This study is the first to address the challenging problem of applying an intentional impact with a kinematic-controlled humanoid robot on non-coplanar contacts.

\end{abstract}

\IEEEpeerreviewmaketitle

\section{Introduction}
% \ak{First since the paper is on balance in multi-contact, you should first quickly cite most pertinent work in this domain, see the RA-L paper of Saeid where there is a good coverage of the topic}

\begin{figure}[!tbp]
  \centering
  \includegraphics[width=0.45\textwidth]{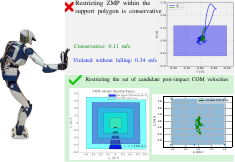}
\caption{The peak impulsive force is 130\unitForce according to the ATI sensor.
  Restricting the ZMP strictly within the support polygon leads to a conservative contact velocity of $0.11$\unitVelTS (see video at \url{https://youtu.be/TL34EWORwbU}).
  Violating the ZMP balance criteria by increasing the contact velocity to  $0.345$\unitVelTS did not result in the robot falling (see video at \url{https://youtu.be/FovQG6U448Q}).
  The robot maintained balance during the experiments because the post-impact CoM velocities did not violate the zero-step capture region, located in the lower-right corner.
  Hence, to determine the maximum contact velocity, we propose restricting the entire set of candidate post-impact CoM velocities $\picomdSet$ (shown as the blue polyhedron in the lower-left corner) within the zero-step capture region. Increasing the region's size would lead to a higher contact velocity, as illustrated by the colored areas and the corresponding  $\picomdSet$ sets.
  %(It's important to note that in our experiments, once the collision was detected, the robot pulled back the impacting end-effector without establising a new sustained contact or exerting additional impulse.)
}
\label{fig:example_one}
\vspace{-3mm}
\end{figure}

\IEEEPARstart{B}{alancing} under multiple, non-coplanar contacts is paramount for humanoid robots to shift weights, redistribute forces, and perform a wider range of tasks in real-world applications, e.g.,~\cite{kheddar2019ram}.
Since many decades, constraining the so-called Zero-tilting Moment Point (ZMP)~\cite{vukobratovic2006ijhr} within the support polygon is considered as an excellent dynamic balancing criteria when the robot stands on a flat surface, i.e., coplanar contacts. Recently, it has been extended to non-coplanar contact in~\cite{caron2016tro}.
However, we experimentally discovered that the ZMP-based criteria can  be temporarily violated without necessarily resulting in loss of balance.
As an example, the impact at the right palm in~\figRef{fig:example_one} enabled the ZMP jumped outside of the support polygon for a while, yet without leading to a fall.

%AK> the whole following paragraph I added is out of scope, it lowers the reading flow, maybe it can be put elsewhere later...
%In~\cite{abi-faraji2019ral}, balance supporting contacts are decoupled from other contact tasks (typically manipulating contacts); whereas in~\cite{samadi2021ral} all contacts (including tasks ones) contribute to balancing. Clearly, an instant contact, namely that resulting from an impact task, shall not be considered as contributing to the balance as it does not sustain the contact.
It is challenging to predict the sudden change of ZMP induced by impact, even with knowledge of the contact velocity and location.
Online implementation of robot control or planning call for low computationally-demanding impact models such as those based on rigid-body dynamics, e.g.,~\cite{zheng1985jfr}. 
These models can predict impact-induced change of velocities, not forces.
As the ZMP is a measure of the resultant force,
predicting its instantaneous change after an impact remains difficult.

As an alternative to ZMP,~\cite{sugihara2009icra} and~\cite{stephens2011thesis} abstracted the robot dynamics with the linear inverted pendulum model (LIPM), and derived falling conditions according to the phase-plane analysis of the center-of-mass (CoM) dynamics.
Their analysis is equivalent to constraining the CoM velocity by the zero-step capture region~\cite{koolen2012ijrr}.
%% Desipte these methods are limited to coplanar contacts,
The impact in \figRef{fig:example_one}, which leads to a temporary violation of the ZMP-based criteria, did not enable the CoM velocity to cross the zero-step capture region. Thus, the comparison suggests that restricting CoM velocity can enable much less-conservative impacts than ZMP.
 \IEEEpubidadjcol
Hence, we establish balance criteria employing CoM velocity to (1) maximize the intentional impact velocity; and (2) avoid the limitations of predicting post-impact ZMP.
With respect to the problem defined in \secRef{sec:problem}, we summarize our contributions from two perspectives:

\emph{Handling non-coplanar contacts:}
In~\secRef{sec:comd_area}, we extend the zero-step capture region approach~\cite{sugihara2009icra,stephens2011thesis,koolen2012ijrr} to include non-coplanar contacts.
To achieve this, we  follow the derivation of the ZMP support area~\cite{caron2016tro} and transform boundaries on ZMP to CoM velocities under the LIPM assumptions.
Next, We project the high-dimensional CoM velocities (represented in the  space of sustained contacts' wrenches) onto a two-dimensional tangent plane of the inertial frame, while satisfying various constraints such as joint torque and velocity limits. The projection follows the ray-shooting algorithm presented in~\cite{bretl2008tro} and its 3D extension in~\cite{audren2018tro}.

\emph{Optimizing the entire set of  post-impact CoM velocities:}
We assume that the robot is kinematic-controlled~\cite{wang2022ral} with high stiffness in joint position or velocity, and that the impact is an instantaneous event occurring over a few milliseconds~\cite{wang2022icra}.
% The second assumption requires the robot to timely detect the collision and pull back the end-effector without exerting  additional impulses~\cite{wang2020ijrr}. 
% We calculate the equivalent momentum at the contact point as if the kinematic-controlled\footnote{Most commercially-available robots are joint velocity (or position) controlled; we refer to them as kinematic-controlled robots.} robot is a composite-rigid-body \cite{orin2013auro,wang2022icra,wang2022ral}.
%% As the impact event fulfills the Coulomb's friction law \cite{stronge2000book,jia2019ijrr} and the principle of momentum conservation,
% With these assumptions,
To approximate the set of candidate CoM velocities, we use convex polyhedra to be aware of all possible post-impact states, as shown in~\figRef{fig:example_one}. By employing the vertices of the polyhedra as optimization variables, we present in~\secRef{sec:contactvel} how to regulate the contact velocity with respect to the predicted sets of post-impact states.

Our study presents a solution enabling a kinematic-controlled legged robot to intentionally impact without losing balance.
Our push-recovery experiment using the full-size humanoid robot HRP-4, employing two non-coplanar contacts on a 30-degree ramp and the ground, validates the zero-step capture region for non-coplanar contacts.
By presenting an optimization-based approach to solve for maximum contact velocities for various HRP-4 stances, including kicking and pushing with non-coplanar contacts, our approach can be easily integrated into other whole-body controllers or planners, enabling legged robots to operate safely without loosing balance at impacts.

\section{Related Work}

% We focus on
% the standing stability margin
% to generate the maximum feasible impact motion without falling.
% This is different from the control design which brings the CoM states to an equilibrium, e.g., the regulator by  \cite{sugihara2009icra} or the hybrid control approach applied by the passive dynamics community \cite{reher2016icra}.

% We compare oursevles to the ZMP-based criteria, and present how we extend the  capture region in \secRef{sec:zmp_dcm}. Impulse is common for robots executing steps. We present the potential to exploit the  proposed criteria for walking in \secRef{sec:walking_impact}.
% We detail how the CoM velocity area  $\comdArea$ was developed from the state-of-the-art in \secRef{sec:multi_contact_area}, i.e., project specifications in higher dimensional space to a 
% low dimensional space following
% \citet{bretl2008tro,caron2016tro}.
This section provides a brief overview of the relevant literature.
In  \secRef{sec:preview}, we discuss existing impact models in robotics.
In  \secRef{sec:zmp_dcm}, we review the capture region and its extensions.
Finally, in \secRef{sec:walking_impact}, we discuss research on on-purpose impact tasks.
%% Specifically, we discuss existing impact models in robotics in \secRef{sec:preview}, as well as the balance capture region and its extensions in \secRef{sec:zmp_dcm}. Additionally, we review research on on-purpose impact tasks in \secRef{sec:walking_impact}.

%% This is a brief review of the knowledge we use in this work. Robotics' impact models in~\secRef{sec:preview}; the balance capture region and extensions in~\secRef{sec:zmp_dcm}; and on-purpose impact tasks in~\secRef{sec:walking_impact}.

\subsection{Impact models}
\label{sec:preview}

% The impact problem determines the the post-impact velocities of a pair of rigid bodies from their pre-impact velocities. It is 
% underconstrained by the momentum conservation. Hence, hypothesis of 
% the physical process are needed, i.e., 
% the Newton’s, Poisson’s, and the Stronge’s restitution laws that are
% defined with respect to (w.r.t.) the velocity, impulse, and energy respectively.

% The state-of-the-art robot control and planning methods are model-based. Whereas,
Several approaches have been proposed to model robotic impacts. In the late 1980s,~\cite{zheng1985jfr} proposed modeling robotic impacts using the algebraic model
based on Newton's law of restitution and assumed frictionless contact.
Model-based controllers or planners can seamlessly integrate this analytical model, according to~\cite{siciliano2016springer,rijnen2019acc}.
% The algebraic-approaches-based linear complementarity problem (LCP)
% is widely used in robotics and computer graphics. However, 
% when the impact is frictional,
% its  solution is not unique as reported by \cite{brogliato1999nonsmooth}.
% Without friction, Poisson’s hypothesis is equivalent to Newton’s hypothesis. However, both of them are energetic inconsistent in the presence of Coulomb’s friction law, see \citet{jia2019ijrr}.

% To name a few examples, see \citet{glocker1995multiple,stewart2000rigid}.

% or resorting to the use of two restitution parameters by \citet{chatterjee1998new}, or the simulated locomotion example by \citet{addi2010impact}, where  two independent, hence decoupled,  LCP problems are solved for the tangential direction contact constraints. The issue is applying some {\bf ratio} of tangential impulse to normal impulse is  theoretically unjustified, see \citet{smith1991predicting}.
% Treating impact dynamics  independently along two tangential directions and normal directions would violate the energy conservation principle, see \cite{jia2017ijrr}.

% The complementarity dynamical system (CDS) by \citet{hurmuzlu2004automatica} offers a general conceptual tool to describe the dynamics of a walking robot. However, CDS applies a similar inelastic frictionless impact model as \cite{zheng1985mathematical}. More importantly, there does not exist an
% effective control design for CDs, see the summary for bipedal walking by \citet{grizzle2014automatica}, and

When the frictional impact is planar (two-dimensional), one can analytically compute the impulse by visiting the intersection points of  the \emph{line of compression}, the \emph{friction cone}, and the \emph{two sliding directions}.
This strategy is known as Routh's graphical approach~\cite{routh1955dynamics} and is considered state-of-the-art impact mechanics~\cite{jia2019ijrr,lankarani2000poisson,khulief2013modeling,wang1992jam}.

Impacts, can be described in 3D similarly to contact forces. In 3D, it is impossible to determine the post-impact tangential velocity, hence the tangential impulse, without numerical integrations~\cite{stronge2000book,jia2017ijrr}.
To the best of the author's knowledge, frictional impact models in 3D are not well-validated for kinematic-controlled robots~\cite{wang2022icra,halm2021ijrr}.
Additionally, incorporating a numerical process (the impact model) into another numerical process (task-space optimization-based controller) is neither straightforward nor computationally efficient in terms of performance and robustness. 

% Meanwhile, LCP does not capture the physical process's progressive nature.
% Recently, following \citet{routh1955dynamics}'s graphical approach, \citet{lankarani2000poisson,jia2017ijrr,jia2019ijrr,halm2019rss}  compute the net impulses and the post-impact states with respect to the evolution of the tangential and normal contact velocities.
% Whereas, these models either assume impacts between free-flying objects \cite{jia2017ijrr,jia2019ijrr,halm2019rss}, or ignore the actuated joints, e.g., the  planar model reported by \citet{lankarani2000poisson} was only validated  against under-actuated pedulums. 

% We adopt \citet{jia2019ijrr} for the solution's existence, uniqueness, and fast computation. Despite  \cite{jia2019ijrr} is restricted to 2D, it is enough for the LIPM model, which assumes constant CoM height. Otherwise, the 3D impact solution is ambiguous for legged robots \cite{remy2017ijrr} or multiple simultaneous impacts \cite{halm2019rss}.

% Accurately capture the impact process's progressive nature requires numerical integration following the velocity-to-impulse mapping at the contact point, see examples by \citet{jia2017ijrr} and \citet{stronge2000book}.     % well-identified models, i.e., the velocity-to-impulse mapping, the restitution and friction coefficients

% as the intersection of the Coulumb's friction cone, the plane of compression, and the plane of restitution.       

% List the references we
% Single ...

% Avoid using the impact duration. 

% Ambiguities....
\subsection{Balance}
\label{sec:zmp_dcm}
\subsubsection{Coplanar contacts}
Restricting the ZMP within the support polygon has been widely used for legged robots walking on flat terrains~\cite{grizzle2014automatica}.
The ZMP balance criteria was also used in impact motions, e.g., hammering a nail~\cite{tsujita2008humanoid}, breaking a wooden piece with a Karate motion~\cite{konno2011ijrr}, or evaluating push-recovery motions~\cite{yi2012iros,rijnen2017icra}. % ; see \cite{wieber2016handbook}.
In the experiments conducted by~\cite{tsujita2008humanoid,konno2011ijrr} the ZMP did not jump outside the support polygon. Thus, 
the impulses exerted by~\cite{tsujita2008humanoid,konno2011ijrr} are not comparable to the example shown in~\figRef{fig:example_one}.

ZMP amounts to a force measurement. Thus, predicting the instantaneous ZMP jump requires accurate impulse (momentum) prediction and proper estimation of the impact duration (time). Impact mechanics based on rigid-body dynamics can predict the impulse.
Whereas, estimating the impact duration is not an easy task, due to the numerous unknown parameters, e.g., the Hertz contact stiffness~\cite[Sec.~4.1]{pashah2008prediction}. Therefore, in this paper, we rather favor an impact-aware balance criteria based on the CoM velocity.

If the divergent component of motion (DCM)~\cite{wieber2016handbook}
% diverges away from the ZMP:
% where we used the pedumlum constant $\omega = \sqrt{\frac{g}{\com_z}}$.
% Through a trivial reformulation of \eqref{eq:dcm_def}
% $
% \comd = - \omega (\com - \dcm),
% $
% we can easily tell that CoM converges to the DCM. Hence the DCM is also referred as the extrapolated CoM \cite{hof2005xcom}.
remains within the support polygon, the CoM stops and stay motionless. Thus the DCM is seen as the capture point~\cite{koolen2012ijrr}.
In~\cite{sugihara2009icra} and~\cite{stephens2011thesis}, the post-impact DCM is restricted within the support polygon, i.e., the \emph{capture region}~\cite{koolen2012ijrr}.
% ; see more details in \remarkRef{remark:dcm}.
% is the same as $\nextStatePostImpact{\state} \in \ssRegion$, 
% When it is allowed to modify the support polygon by taking one or more steps, we can analyze the capturability, or essentially the viability, by the  .
In this paper, we consider a fixed stance during impact, i.e., without modifying the sustained contacts. Hence, it is not comparable to the \emph{N-step capture region} discussed in~\cite{koolen2012ijrr,posa2017rss}. 

% or the seminal push-recovery MPC (PR-MPC) by \cite{stephens2010humanoids} which updates foot location and the CoM trajectory.
% The PR-MPC is different from the Q.~\ref{item:q-best-contact} since it does not preview the post-impact CoM velocities and it assumes a biped robot with coplanar contacts.
% In the thesis by Stephens~\cite{stephens2011thesis}, taking a step or not (applying the PR-MPC), is determined by the capture region. Thus the PR-MPC is a controller that the proposed standing stability margin can not compare with.

% trajectory of a pre-defined coplanar foot contact. 

% , we add additional objectives to the
% push recovery MPC (PR-MPC) to find the modified foot location, i.e., { I.\ref{improvement:foot-step}}, such that bounds of
% the controllable region, i.e., bounds of the CoM velocity, is maximized with respect to the predicted impact-induced CoM velocity jump.

% (*) Impact in walking how to judge the Stability

% (*) The multi-contact paper by Stephans
\subsubsection{Non-coplanar contacts}
\label{sec:multi_contact_area}
% The impact-unaware standing stability criteria are well-established. We noticed

%% In~ the capturability criteria is proposed with non-coplanar yet not computing the entire

We define
the multi-contact wrench cone (MCWC)
% the set of possible resultant wrenches
at the robot CoM as the 
\href{https://www8.cs.umu.se/kurser/TDBAfl/VT06/algorithms/BOOK/BOOK5/NODE199.HTM}{Minkowski sum} of the individual stable contact wrench cone (CWC)~\cite{caron2015icra}.
% we compute the 
% \href{https://www8.cs.umu.se/kurser/TDBAfl/VT06/algorithms/BOOK/BOOK5/NODE199.HTM}{Minkowski addition}
% of CWCs to obtain
% the multi-contact wrench cone (MCWC), 
Employing the relation between ZMP and the resultant wrench,~\cite{caron2015icra} projects the MCWC on a desired, two-dimensional plane to obtain the feasible ZMP support area. 
The projection follows the ray-shooting algorithm~\cite{bretl2008tro,audren2018tro}, a simpler yet conservative approach is proposed in~\cite{samadi2021ral}. 

Given (1) MCWC, and (2) the relation between CoM velocity and ZMP, we can compute the CoM velocity area  $\comdArea$. As long as the post-impact CoM velocity belongs to $\comdArea$, 
the sustained contacts can balance the robot. 

Calculating the ZMP area $\zmpArea$ assuming unlimited contact wrenches is not realistic, as it ignores kinematic and dynamic limitations.
Hence,~\cite{caron2016tro} reduced $\zmpArea$ by making the LIPM assumptions, which include restricting  the derivative of angular momentum to zero, maitaining a fixed CoM height, and fixing gravitational direction acceleration. More recently,~\cite{orsolino2019tro} further improved $\zmpArea$ by considering joint torque limits.

Note that we compute the full zero-step capture region rather than checking the zero-step capturability of a given stance \cite{del2018tro}, i.e., predicting the fall given the particular contact configurations, CoM position and velocity without computing the boundaries of the zero-step capture region.
%% are essential for a whole-body controller to find the contact velocity without leading to a fall.

\subsection{On-purpose impact tasks}
\label{sec:walking_impact}
% \subsubsection{Stability criterion for walking}
Dynamic walking frequently exerts impacts as impact-less reference trajectories are challenging to generate and inefficient to execute~\cite{grizzle2014automatica}.
State-of-the-art ZMP or DCM based walking control design ignores impact dynamics, e.g., see the examples by~\cite{kajita2010iros,feng2015optimization}.
The proposed criteria can enable  far-less conservative impact motion, e.g., kicking velocities.
Therefore, we can complement a whole-body controller~\cite{koolen2016ijhr}, or a planner \cite{gao2020acc}, to generate impact motions
or balance behaviors that drastically change the centroidal momenta in a short time.
% For example, the admissible ZMP jump is considerably higher than usual as shown in \figRef{fig:example_one}.

%% Thus, it is has the potential to improve the reference trajectory generation.
% we can  enable simpler computation to generate  more  dynamic motion.

% Alternative to applying a passive dynamics strategy, e.g., \cite{reher2016icra},

\section{Problem Formulation}
\label{sec:problem}
We formulate our research problem with three steps.
First,  in \secRef{sec:basics}
we introduce the mathematical tools to be used throughout the paper. 
Second, in \secRef{sec:analysis} we use the phase portrait to analyze why the impact experiments shown in~\figRef{fig:example_one} did not lead to a fall.
Then, in~\secRef{sec:problem} we briefly state the research problem.

\subsection{Mathematical preliminaries}
\label{sec:basics}
We will define the ZMP, the relationship between the ZMP and the CoM acceleration, and the linear inverted pendulum model (LIPM).

\begin{figure}[!htbp]
  \centering
  \includegraphics[width=0.45\textwidth]{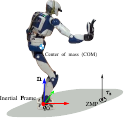}
  \caption{
    At the ZMP, the torque (black arrow) of the resultant contact wrench $\torque_{\zmp}$ aligns with the normal vector $\nz$, which is the Z axis (blue arrow) of the inertial frame. }
\label{fig:frames}
\end{figure}

% Linear and angular momentum
%% Suppose there exist $m$ sustained contacts between the robot and the environment,

In \figRef{fig:frames}, we define the inertial frame $\cframe{\inertialFrame}$ whose origin $\point{\inertialFrame}$ locates under the CoM, i.e., $\comPlane = [0, 0]^\top$.

Suppose the robot has $\numContacts$ sustained contacts, we align the orientations of all the forces and wrenches according to the inertial frame. 
  For instance,  if the $i$-th contact wrench $\wrenchFrame{i}{i}$ is represented in the local frame, we represent $\wrenchFrame{i}{i}$ using the inertial frame's orientation as:
$$
\wrench_{i}
=
\matrixTwo{\rotation{i}{\inertialFrame}}{0}{0}{\rotation{i}{\inertialFrame}}
\wrenchFrame{i}{i}
$$
where $\rotation{i}{\inertialFrame}$ denotes the rotation from
the $i$th contact frame $\cframe{i}$ to the inertial frame $\cframe{\inertialFrame}$.
%% the following quantities
%% to the orientation of the inertial frame $\cframe{\inertialFrame}$: (1) the contact forces and locations  $\force_{i} \in \RRv{3}$, $\contactPoint_i \in \RR^3$  for $i = 1,\ldots, \numContacts$, (2) the linear and angular momentum: $\linearMomentum \in \RRv{3}$,  $\angularMomentum \in \RRv{3}$.

%% \begin{remark}
  
%%  \backfill
%% \end{remark}

Hence, we can compute the resultant wrench $\inertialWrench$ at the  origin of the inertial frame $\point{\inertialFrame}$ as:
%% $\wrench_{\inertialFrame} = \vectorTwoRow{\force_{\inertialFrame}}{\torque_{\inertialFrame}}$
%% To constrain  $\wrench$, we substitute the following summation in  \eqref{eq:angular_momentum_assumption_constraint}:
\quickEq{eq:sum_wrench}{
\inertialWrench = \wrenchC{\inertialForce}{\inertialTorque} = \idMatrix{\wrench} \wrench =
\underbrace{
  \begin{bmatrix}
    \idMatrix{1} &\idMatrix{2} & \ldots & \idMatrix{\numContacts}
  \end{bmatrix}
}_{\idMatrix{\wrench} \in \RRm{6}{6\numContacts}}
  \underbrace{\begin{bmatrix}
\wrench_1 \\ 
\vdots\\ 
\wrench_{\numContacts}
\end{bmatrix}}_{\wrench  \in \RRm{6\numContacts}{1} }.
}
The matrix $\idMatrix{\wrench}$ horizontally collects $\numContacts$ blocks: 
$$
\idMatrix{i} = \matrixTwo{\identityMatrix}{\zeroMatrix}{\pointer{\point{\inertialFrame}}{\contactPoint_i} \times}{\identityMatrix} \in \RRm{6}{6}, \quad \text{for}\quad i = 1, \ldots, \numContacts,
$$
where $\identityMatrix \in \RRm{3}{3}$ denotes the identity matrix.
The vector $\pointer{\point{\inertialFrame}}{\contactPoint_i} = \pointerDef{\point{\inertialFrame}}{\contactPoint_i}$ connects a contact point $\contactPoint_i$ to the
origin $\point{\inertialFrame}$ of the inertial frame.
Thus, $\pointer{\point{\inertialFrame}}{\contactPoint_i} \times $ computes  the resultant torque for a given sustained contact's force.
%% $$
%% \geometricFT{e}{\com} = \geometricFTDef{e}{\com} = \geometricFTTwoDef{e}{\com}
%% $$

\subsubsection{ZMP}
Given a unit vector $\nz$, the Zero-tilting Moment Point (ZMP) \cite{vukobratovic2006ijhr,caron2016tro,sardain2004tro} denotes the point where the torque of the resultant contact wrench $\torque_{\zmp}$ aligns with $\nz$, see \figRef{fig:frames}:
\quickEq{eq:zmp_condition}{
\nz \times \torque_{\zmp} = 0.
}
Computing the resultant wrench and the expansion of the vector triple product,
we can define the ZMP as: 
\begin{equation}
  \label{eq:def_zmp}
  \zmp = \frac{\nz \times \inertialTorque}{\innerP{\nz}{\inertialForce}}  +  d_{\zmp} \frac{\inertialForce}{\innerP{\nz}{\inertialForce}},
\end{equation}
where we leave the detailed derivaiton in \appRef{app:cwc}.
If we assume that the scalar $d_{\zmp} = \innerP{\bs{n}}{\zmp} = 0$, which means  $\zmp$ belongs to the
surface
%% $\plane{O}{\bs{n}}$
with the origin $\point{\inertialFrame}$ and the surface normal $\bs{n}$, we have the simpler definition: 
\quickEq{eq:zmp_def_simple}{
  \zmp = \frac{\nz \times \inertialTorque}{\innerP{\nz}{\inertialForce}}.
}
\subsubsection{CoM acceleration and ZMP}
According to the derivation in \appRef{app:com-zmp}, the CoM acceleration $\comdd$ depends on the ZMP $\zmp$ and angular momentum's derivative $\angularMomentumD$:
\begin{equation}
\label{eq:three_relation}
\comdd = \gForce + \frac{\cross{\dot{\angularMomentum}}{\nz}}{\mass (\height{\com} - \height{\zmp})} 
+ \frac{\innerP{\nz}{( \comdd - \gForce)} \pointer{\com}{\zmp}}{ 
\height{\com} - \height{\zmp}}.
\end{equation}
\subsubsection{The LIPM model}
The LIPM model  \cite{sugihara2009icra,caron2019icra} simplifies the whole-body dynamics \eqref{eq:three_relation} with two assumptions:
\begin{enumerate}
\item The vertical CoM acceleration is zero:
\begin{equation}
\label{eq:lipm_ass_1}
\innerP{\nz}{\comdd} = 0. 
\end{equation}
\item The angular momentum about the CoM is fixed: 
\begin{equation}
\label{eq:lipm_ass_2}
\dot{\angularMomentum} = 0.
\end{equation}
\end{enumerate}
Substituting  \eqref{eq:lipm_ass_1} and \eqref{eq:lipm_ass_2}, we can simplify  \eqref{eq:three_relation} to:
\quickEq{eq:lipm-model-3d}{
\comdd = \gForce  + \frac{g}{\height{\com} - \height{\zmp}} (\com - \zmp),
}
where the detailed derivation is left in  \appRef{app:lipm-model-3d}. 
If we project ZMP on the ground surface, i.e., $\height{\zmp}= 0$, and define the pendulum constant
$\pendulumC = \pendulumCDef$,  the planar components of \eqref{eq:lipm-model-3d} writes:
\quickEq{eq:lipm-model}{
\comdd = \pendulumC^2(\com - \zmp).
}

% The relation
\subsection{Problem analysis}
\label{sec:analysis}
\begin{figure}[htbp!]
  \includegraphics[width=0.48\textwidth]{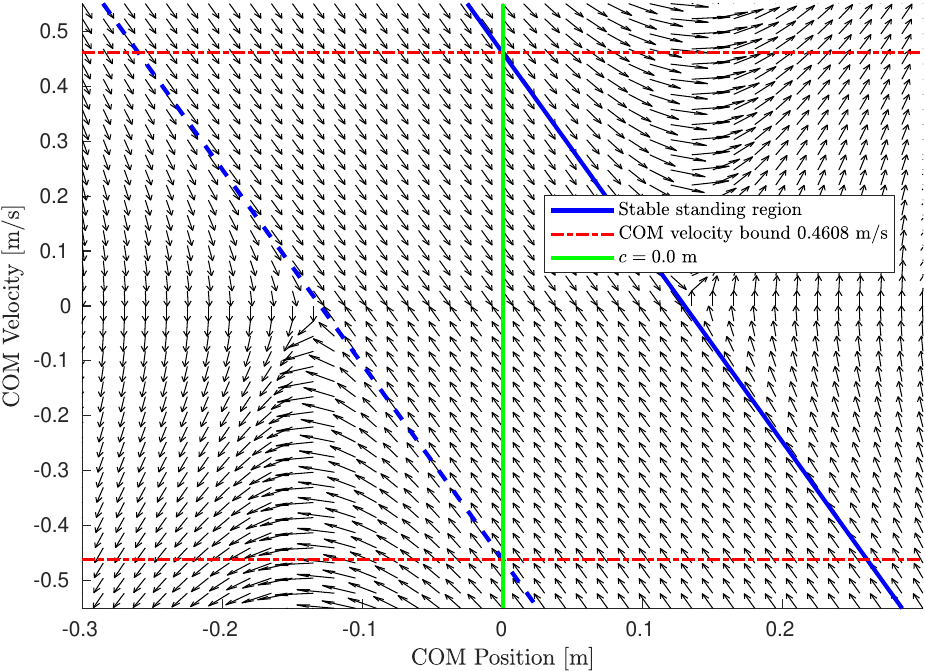}
  \caption{
    Simulation of the phase portrait of the LIPM dynamics \eqref{eq:ss-lipm-model} for the HRP-4 stance in Fig.~\ref{fig:example_one}, following \cite{sugihara2009icra}. The ZMP is saturated within the interval $[-0.13, 0.13]$\unitPos, and the CoM height is $0.78$\unitPos.
  }
  \label{fig:lipm-pplane}
\end{figure}

% \todo{Visualize the coordinate frames: inertial frame $\cframe{\inertialFrame}$, centroidal frame $\cframe{\com}$, contact frame $\cframe{\contactPoint}$, and the instantaneous inertial frame $\cframe{\iiFrame}$ (whose origin locates on the ground and shares the CoM's X-Y coordinates).
% }
%% as:
%% \begin{equation}
%%   \label{eq:com_dcm_dynamics}
%%   \vectorTwo{\comVel}{\dcmd}
%%   = \pendulumC \matrixTwo{
%%     -1
%%   }{
%%     1
%%   }{
%%     0
%%   }{
%%     1
%%   }
%%   \vectorTwo{
%%     \com
%%   }{
%%     \dcm
%%   }
%%   +
%%   \vectorTwo{0}{\pendulumC}\zmp,
%% \end{equation}
%% which includes
Choosing the state variable $\state = [\com_x,\comd_x]^\top$, the state-space form of the LIPM dynamics \eqref{eq:lipm-model} along the sagittal direction writes: 
\begin{equation}
\label{eq:ss-lipm-model}
  \vectorTwo{\comd_x}{\comdd_x}
  =
  \matrixTwo{
    0
  }{1}{\pendulumC^2}{0}
  \vectorTwo{
    \com_x
  }{
    \comd_x
  }
  +
  \vectorTwo{0}{-\pendulumC^2}
  \zmp_x.
\end{equation}
The phase portrait of \eqref{eq:ss-lipm-model} in \figRef{fig:lipm-pplane} shows the \emph{stable standing region} \cite{sugihara2009icra,stephens2007humanoid}:
%% \begin{equation}
%%   \label{eq:def_stable_standing_region}
%%   \begin{aligned}
%%   \ssRegion &\setDef{\dcm}{ \ssrLB \leq \dcm \leq \ssrUB },\\
%%    \ssrUB &:\quad \error{x_1} + \frac{\errord{x_1}}{\pendulumC} = \upperBound{\zmp}, \\
%%    \ssrLB &:\quad \error{x_1} + \frac{\errord{x_1}}{\pendulumC} = \lowerBound{\zmp}.
%% \end{aligned}
%% \end{equation}
\begin{equation}
  \label{eq:def_stable_standing_region}
  \ssRegion \setDef{\dcmxDef}{ \lowerBound{\zmp_x} \leq \dcmxDef \leq \upperBound{\zmp_x} },
\end{equation}
where the ZMP is limited by: $\zmp_x \in [\lowerBound{\zmp_x}, \upperBound{\zmp_x}]$.
%% Note that the stable standing region $\ssRegion$ is fixed regardless of the choice of ZMP feedback.
% We leave the proof of \eqref{eq:def_stable_standing_region} in  \appRef{app:ssr-proof}.

For the HRP-4 stance in \figRef{fig:example_one}, we place
the inertial frame's origin $\point{\inertialFrame}$ under the CoM,
i.e., the CoM coordinates is: $\com = [ 0, 0, 0.78]$\unitPos.
Thus, substituting $\com_x = 0$ into \eqref{eq:def_stable_standing_region} simplifies $\ssRegion$ as:
$$
\ssRegion \setDef{\frac{\comVel_x}{\pendulumC}}{ \lowerBound{\zmp_x} \leq \frac{\comVel_x}{\pendulumC} \leq \upperBound{\zmp_x} }.
$$
Given   $\com_z = 0.78$\unitPos, $g = \gValue$\unitAccTS, and $\zmp \in [-0.13, 0.13]\unitPos$, we can find the bounds: $\comVel_x \in [-0.4608, 0.4608]$\unitVelTS. Thus, the two intersection points $[0, 0.4608]$, $[0, -0.4608]$ in \figRef{fig:lipm-pplane} (between the green vertical line and the $\ssRegion$) defines the interval, within which the CoM velocity can converge back to the origin.

\begin{figure}[htbp!]
  \includegraphics[width=0.48\textwidth]{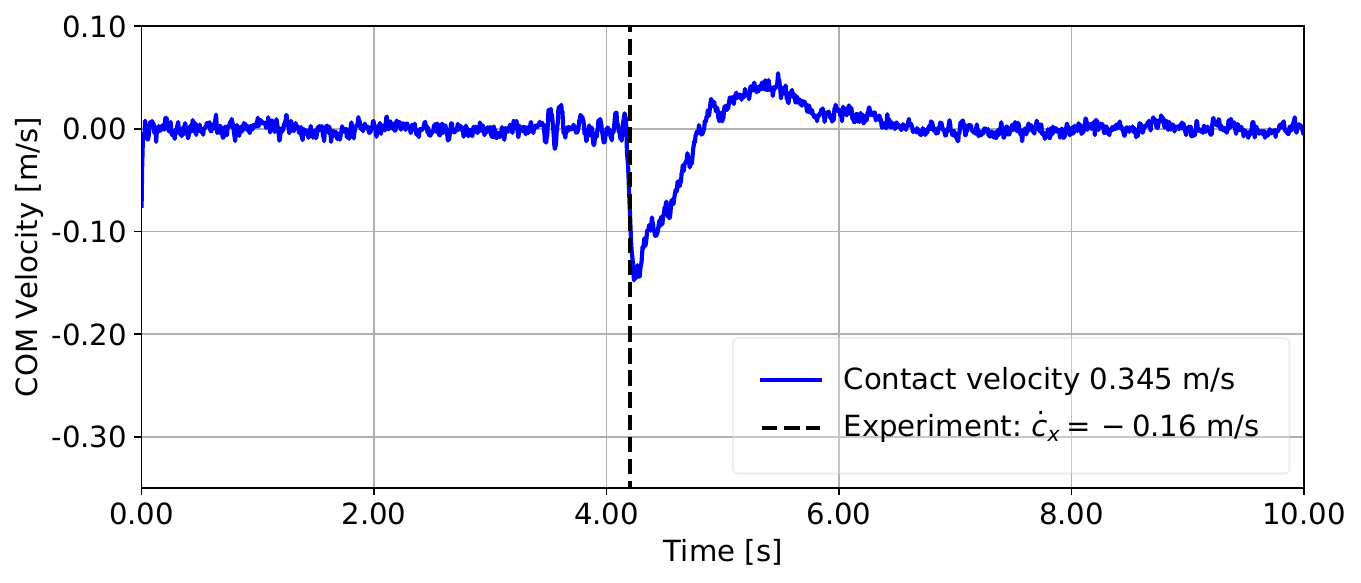}
  \caption{
    The CoM velocity $\comd_x$ at the impact time (indicated by the black dashed line) for the HRP-4 stance in Fig.~\ref{fig:example_one} was within the analytical bound: $-0.16$\unitVelTS $\in [-0.4608, 0.4608]$\unitVelTS.
  }
  \label{fig:com-vel}
\end{figure}

 Thus, despite the contact velocity $0.345$\unitVelTS violated the ZMP-based criteria as shown in \figRef{fig:example_one},
 it cannot cause a fall. Because the post-impact CoM velocity $-0.16$\unitVelTS in \figRef{fig:com-vel} is within the interval $[-04608, 0.4608]$\unitVelTS.

% We plotted both of them in \figRef{fig:pplane-best-com-zmp}, where we can find the convergence region (the \emph{confining region} by \cite{sugihara2009icra}) is defined by
% the intersection of \eqref{eq:def_controllable_region} and \eqref{eq:def_stable_standing_region}.
% % , and the trajectories passing through the intersection.

However,  the phase-portrait analysis in \figRef{fig:lipm-pplane} has limitations: (1) the stable standing region  $\ssRegion$ only applies in one direction, e.g., the sagittal direction; (2) the robot is restricted to coplanar contacts; and (3) intentional impacts are not considered. Hence, we formulate our research problem in the next subsection.

\subsection{Problem statement}
\label{sec:problem}
\begin{problem}
  \label{prob:standing}
  %% In order to define the control problem rigorously,
  We adopt the following assumptions for an intentional impact task:
  \begin{enumerate}
  \item The robot joint configurations $ \jangles \in \RRv{\jsDim}$, $\jvelocities \in \RRv{\jsDim}$ are known (measured or observed);
  \item Prior to the impact, the robot's initial multi-contact configuration is balanced (in theory this condition is conservative and can be relaxed).
  \item The robot is high-stiffness controlled either in joint velocity or position, i.e., kinematic-controlled.
    It should be noted that robots with a different joint-control mode, such as a pure-torque controlled humanoid robot TORO, may behave differently during impact events.
    \item There are $\numContacts$ sustained non-coplanar rigid contacts with known friction coefficients;
    \item We can approximate any of the sustained contacts' geometric shape with a rectangle;
    \item The robot controller can timely detect the collision, and pull back the end-effector without exerting additional impulses.
    \item The impact does not break the sustained contacts during the impact event, which typically lasts for dozens of 40\unitMS in our previous experiments~\cite{wang2022ral} (this condition can indeed be enforced by limiting the impact direction and intensity).
  \end{enumerate}
  %% (5) We also assume the hardware resilience bounds, e.g., joint velocity and torque limits, are known.
  A task planner, or a human operator, typically provides a reference contact relative velocity $\reff{\contactVel} \in \RRv{3}$ for a given end-effector, e.g., the right palm in~\figRef{fig:sample_posture} is asked to impact along the yellow-arrow's direction.
However, when the reference velocity  $\reff{\contactVel}$ is too high it can lead to loosing balance and subsequent fall; the question is how to determine the maximum tracking velocity that does not cause falls.
  \begin{figure}[htbp!]
    \centering
    \includegraphics[width=0.85\columnwidth]{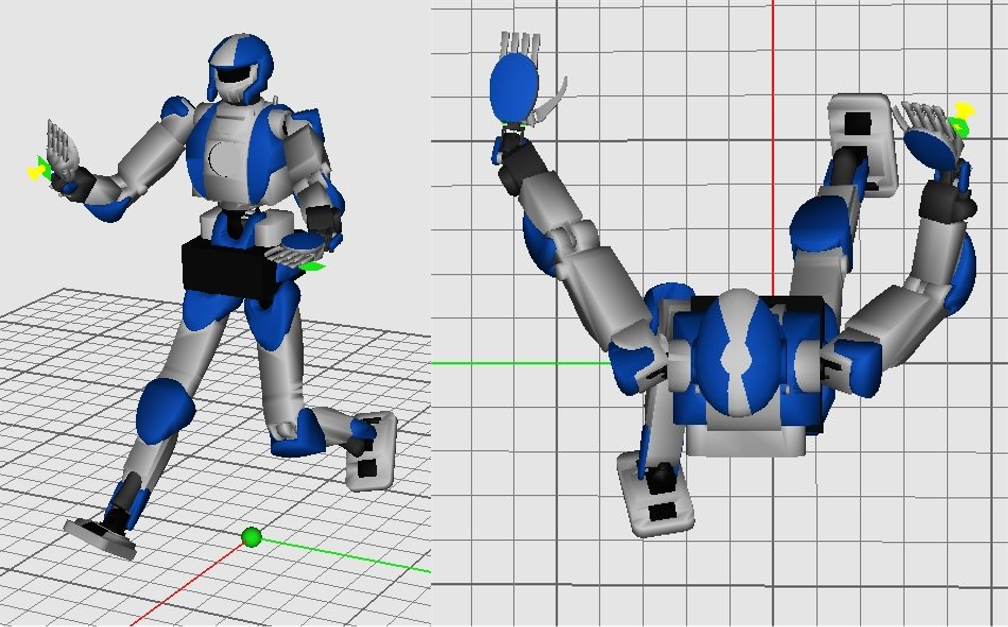}
    \caption{
      Side (left) and top (right) views of the HRP-4 robot with two non-coplanar contacts, i.e., two feet contacts with friction coefficient $0.7$. The robot applies impact with the right palm, as indicated by the yellow arrow.   
    }
    \label{fig:sample_posture}
  \end{figure}
  % ensured via impact-aware task-space control design \cite{wang2020ijrr}.
  \backfill
\end{problem}

%Alternative to enforcing balance using ZMP, which is a measure of force, CoM velocity can been seen as the integral of the resultant forces.
Controllers designed to be impact-aware, such as the one described in \cite{wang2020ijrr}, take into account the frictional impact dynamics.
Being able to predict the post-impact states, these controllers can use an impact event as a controlled process for various purposes, e.g., kicking or hammering.
Therefore, we treat the impact as an instantaneous event that
injects a predictable amount of impulse into the CoM dynamics and address \probRef{prob:standing}  in two steps:
%% As CoM velocity is less conservative than ZMP, i.e., the 0.34\unitVelTS impact  in \figRef{fig:example_one} enabled ZMP momentarily jumped outside the support polygon, yet COM velocity sustained within the capture region.
\begin{enumerate}
\item Given the sustained  contacts and joint configurations, we compute the CoM velocity area $\comdArea$  in \secRef{sec:comd_area}.
  As long as $\picomdxy \in \comdArea$ after the impact, the robot can balance  despite the impulse;
\item In \secRef{sec:contactvel},
  we compute the set of post-impact CoM velocities $\picomdSet$, and  formulate novel inequality constraints to guarantee that $\picomdSet \subseteq \comdArea$.
\end{enumerate}

% In addition to standing stability, impact
% \citet{wang2020ijrr} ensures the resillience bounds. 
% To address the above problem, 

% How to represent an impulse cone? Basis vector and 
% Given the pre-impact robot configurations, friction coefficient, and the restitution coefficient, the
% $\impulse \in \impulseCone$

% $\surfaceNormal \in \normalCone$

\section{The CoM Velocity Area}
\label{sec:comd_area}
%% Given   the sustained contacts' configurations and LIPM assumptions, this section summarizes the CoM velocity area.
%% In order for the robot to maintain balance after the impact, the CoM velocity must fall within a specific range known as the CoM velocity area. This range is defined in Section IV, which is divided into three subsections.

After an impact event, the robot can balance utilizing the sustained contacts as long as the CoM velocity is within a specific range,
which we refer to as the CoM velocity area.
We present the CoM velocity area in three subsections. 
In \secRef{sec:eqm_constraint}, we  summarize the set of resultant wrench $\set{\inertialWrench}$ at the origin of the inertial frame. 
%% A robot controller can regulate/stablize the CoM velocity to the origin with the resultant wrench.   
In \secRef{sec:set-comd}, 
we derive boundaries of the set of CoM velocities $\comdArea$, that can be balanced according to $\set{\inertialWrench}$.
Finally, in \secRef{sec:lp_projection}, we project the high-dimensional-represented set $\comdArea$  onto the two-dimensional tangent plane of a humanoid robot following~\cite{caron2016tro,bretl2008tro}.

%% velocity area .

%% denote the set of CoM velocities that a robot can balance be ba

%% \secRef{sec:eqm_constraint} presents the resultant wrench cone at the CoM given the sustained contacts. 
%% .

% The feasible CoM velocities are defined with respect to the sustained-contact  wrenches in a $6\numContacts$ dimensional space, i.e., the linear constraints (\ref{eq:cwc_def}, \ref{eq:torque_limits}, \ref{eq:angular_momentum_assumption_constraint}, \ref{eq:comd_equality}).

% (\ref{eq:cwc_def}, \ref{eq:torque_limits}, \ref{eq:angular_momentum_assumption_constraint}, \ref{eq:comd_equality}) to a two-dimensional plane.
 % presents the customized projection from a high-dimensional space 

\subsection{The set of resultant wrenches}
\label{sec:eqm_constraint}

Given $\numContacts$ sustained contact wrenches $\wrench \in \RRm{6\numContacts}{1}$ in \secRef{sec:cwc}, we constrain the resultant wrench at the inertial frame's origin $\point{\inertialFrame}$ according to limited actuation torques in \secRef{sec:torque_limit}, and the LIPM assumptions in \secRef{sec:relaxed_LIPM_assumptions}.
% Each wrench 
% fulfills the contact wrench cone (CWC) \eqref{eq:cwc_def} to sustain a stable contact.
% Converting to joint space, $\wrench$ respects the torque limits by the equations of motion \eqref{eq:torque_limits}. Given the Newton-Euler equations, the sum of the resultant wrenches at the CoM
% needs to fulfill 
% \eqref{eq:angular_momentum_assumption_constraint} to meet the relaxed LIPM assumptions (\ref{eq:constant_mg}-\ref{eq:lipm_ass_3}).
\subsubsection{Contact Wrench Cone}
\label{sec:cwc}
A  sustained contact wrench $\wrench_i \in \RRv{6}$ fulfills the Coulumb's friction cone, has a limited torque, and does not slip. Assuming a rectangular contact area,  Caron~\emph{et al.}~\cite{caron2015icra}  collected the three conditions into a half-space representation:
\quickEq{eq:cwc_def}{
  \set{\wrench_i} \setDef{
    \wrench_i \in \RRv{6}
  }{
    \cwc_i \wrench_i \leq 0
  },
  \quad \text{for } i = 1,\ldots,\numContacts,
}
where the details of $\cwc_i \in\RRm{16}{6}$ are available in \appRef{app:cwc}.

\subsubsection{Limited actuation torques}
\label{sec:torque_limit}
According to the Newton's third law,
the sustained contact wrenches $\wrench \in \RRm{6\numContacts}{1}$ are subject to the joint actuator torques limits $ \actuationMatrix \lowerBound{\torque}  \leq \actuationMatrix \upperBound{\torque} \leq \actuationMatrix \upperBound{\torque} $. Thus, we re-formulate the equations of motion to restrict the sustained contacts' wrenches $\wrench$ accordingly:
% We discard blocks corresponding to the floating-base joints from the equations of motion:
% \quickEq{eq:torque_limits_constraints}{
%   \begin{aligned}
%     - \agg{i}{\numContacts}{\transpose{\jacobian_i}\wrench_{i}} & \leq \upperBound{\torque} - \inertiaMatrix \jaccelerations - N \\
%     \agg{i}{\numContacts}{\transpose{\jacobian_i}\wrench_{i}} & \leq   - \lowerBound{\torque} + \inertiaMatrix \jaccelerations + N
%   \end{aligned}, 
% }
% and integrate \eqref{eq:torque_limits_constraints} to \eqref{eq:zmp-projection-inequality} as:
% We leave the discussion about coordinate frame in \remarkRef{remark:frame} and the reason why only the $\numContacts$ sustained contacts are considered in   \eqref{eq:torque_limits_constraints} in \remarkRef{remark:instantaneous}. 
\quickEq{eq:torque_limits}{
\underbrace{
  \vectorTwo{
    -\transpose{\jacobian}
  }{
    \transpose{\jacobian}
  }
}_{A}
\wrench \leq
\underbrace{
  \vectorTwo{
    \actuationMatrix \upperBound{\torque} - \inertiaMatrix \jaccelerations - \gravityandcoriolis
}{
  - \actuationMatrix \lowerBound{\torque} + \inertiaMatrix \jaccelerations + \gravityandcoriolis
}
}_{B}.
}
The detailed definition of $\jacobian, \inertiaMatrix, \gravityandcoriolis, \actuationMatrix$ are left in \eqref{eq:eom} as part of \appRef{app:eom}.

% dynamic robot motion constraint
% \begin{remark}
%   \label{remark:frame}
%   where the matrices including the Jacobian $\jacobian$ are taken from the robot equations of motion and therefore by default represented in the inertial frame $\cframe{\inertialFrame}$. Re-writing the above as constraints on the wrenches:
%   Note that if the wrench $\wrench_{i}$ is not aligned with the inertial frame $\cframe{\inertialFrame}$, we need to rotate $\wrench_{i}$ accordingly. 
% \end{remark}
% \begin{remark}
%   \label{remark:instantaneous}
%   The predicted impact-induced contact force jump is instantaneous, thus it is not
%   accounted for the multi-contact ZMP area $\zmpArea$ calculation, where only sustained contact wrenches are considered.

%   Despite the instantaneous modification of the CoM velocity, as long as \eqref{eq:stability_condition} is fulfilled, the robot will not fall, see the phase-portraits in \figRef{fig:pplane-best-com-zmp}.
%   \backfill
% \end{remark}

\subsubsection{LIPM Assumptions}
\label{sec:relaxed_LIPM_assumptions}
%% \todo{
%%   Do we need to project the sustained contact wrenches to the CoM here?
%% }
To enforce the LIPM assumptions~\eqref{eq:lipm_ass_1} and~\eqref{eq:lipm_ass_2}, we choose to constrain the resultant wrench~$\inertialWrench = \vectorTwoRow{\inertialForce}{\inertialTorque}$.
\begin{enumerate}
\item For the \emph{zero vertical CoM acceleration} assumption \eqref{eq:lipm_ass_1}, we restrict the resultant force $\inertialForce$:
\begin{equation}
\label{eq:constant_mg}
\innerP{\nz}{\inertialForce} = \mass g,
\end{equation}
where we  use the scalar $\mass g$ instead of $\mass \gForce$.
The gravity $\gForce$ takes the opposite sign of the $z$-axis of the inertial frame $\nz = [0, 0, 1]^\top$, i.e., $\innerP{\bs{g}}{\nz} = -\gValue$.
\item For the \emph{fixed angular momentum} assumption \eqref{eq:lipm_ass_2},
  we restrict the resultant torque at the CoM  $\torque_{\comFrame}$ according to \cite{caron2016tro}:
  %%   % Different from \cite{caron2016tro,orsolino2019tro},
  %% \todo{Derivative of the angular momentum ${\angularMomentumD}_{\com}$ is zero}:
\begin{enumerate}
\item Parallel to the Z axis $\nz =   [0, 0, 1]^\top$ of the inertial frame $\cframe{\inertialFrame}$:
  %% Torque about the center of mass $\torque_{\com}$:
\begin{equation}
\label{eq:lipm_ass_3}
\cross{\nz}{\torque_{\comFrame}} = 0.
\end{equation}
\item Zero torque about $\nz$:
\begin{equation}
\label{eq:assumption_1}
\innerP{\nz}{\torque_{\comFrame}} = 0.
\end{equation}
\end{enumerate}
\end{enumerate}

We re-formulate (\ref{eq:constant_mg}-\ref{eq:assumption_1}) in a matrix form \eqref{eq:angular_momentum_assumption_constraint}. The detailed steps are left in \appRef{app:com-moment}.
\begin{equation}
  \label{eq:angular_momentum_assumption_constraint}
\begin{bmatrix} 
\height{\com} \cdot \identityMatrix &\cross{\nz}{}\\
-(\cross{\nz}{\com})^\top & \nz^\top 
\end{bmatrix}
\wrenchC{\inertialForce}{\inertialTorque} = 
% \underbrace{\wrenchC{\force}{\torque_{O}}}_{\wrench} =
\begin{bmatrix}
\mass g\com\\
0
\end{bmatrix}.
\end{equation}
%% where $\force_{\inertialFrame}, \torque_{\inertialFrame} \in \RRv{3}$ denote the sum of the external forces and the resultant torque at the inertial frame origin  $\point{\inertialFrame}$.
Thus, we summarize the set of resultant wrenches at the inertial frame's origin as:
\quickEq{eq:resultant_wrenches}{
  \set{\inertialWrench}  \setDef{
    \inertialWrench \in \RRv{6}
  }{
    \inertialWrench \text{~fulfills \eqref{eq:angular_momentum_assumption_constraint}
      %% \eqref{eq:torque_limits}
    }
  }.
}

%% \todo{
%%   The equality \eqref{eq:angular_momentum_assumption_constraint} limits the contact wrenches $\wrench \in \RRv{6\numContacts}$ of $\numContacts$ sustained contacts according to the LIPM assumptions.
%% }

%% \todo{
%%   The constraints \eqref{eq:torque_limits}\eqref{eq:angular_momentum_assumption_constraint}
%%   presents the set of feasible resultant wrenches at the CoM. 
%% }

% where the wrench $\wrenchFrame{i}{i}$ is measured and represented at the local frame $\cframe{i}$\footnote{Note that the frame  $\cframe{i}$ is aligned with the inertial frame $\cframe{\inertialFrame}$, namely, the relative rotation is identity: $\rotation{i}{\inertialFrame} = \identityMatrix$.} and  $I$ is 
%  is a  horizontal stack of 6 dimensional identity matrices
% \begin{equation}
% \label{eq:def_I}
% I = \begin{bmatrix}
%  I &I & \ldots & I
% \end{bmatrix}.
% \end{equation}

\subsection{
  Boundaries of the  balancable CoM velocity set}
\label{sec:set-comd}
We denote the set of CoM velocities $\comdArea$, within which the robot can stablize the CoM with the sustained contacts, i.e., the set of resultant wrenches  $\set{\inertialWrench}$.

By fixing the inertial frame's origin  $\point{\inertialFrame}$ under the CoM, we
can simplify the stable standing region $\ssRegion$ by substituting $\comPlane = [0, 0]^\top$ into \eqref{eq:def_stable_standing_region}:
$$
\ssRegion \setDef{\comdPlane}{ \pendulumC\lowerBound{\zmp} \leq \comdPlane \leq \pendulumC\upperBound{\zmp} }.
$$
Thus, at the boundary of the set $\set{\comdxy}$, the planar CoM velocity $\comdxy$ and ZMP $\zmp \in \RRv{2}$ fulfill:
\quickEq{eq:comd_zmp}{
  \comdxy = \pendulumC \zmp.
  %% , \text{where}~ c_1 =
  %% \begin{dcases*}
  %%   -\frac{\pendulumC}{ \pole{2} - 1}   & \text{strong standing stability}\\
  %%   \pendulumC & \text{weak standing stability},
  %% \end{dcases*}
}

We can write ZMP as a function of the CoM and $\inertialForce$, see \appRef{app:zmp-com-force}, 
\quickEq{eq:mc_zmp_projection}{
  \zmp = \comPlane + \frac{\height{\zmp} - \height{\com}}{\mass g}\inertialForce. % + \frac{1}{mg}\cross{\bs{n}}{{\angularMomentumD}_{\com}}.
}
% The detailed  derivation of \eqref{eq:angular_momentum_assumption_constraint} and \eqref{eq:mc_zmp_projection} are left in \appRef{sec:mc-zmp-area-computation}.
substituting $\comPlane = [0, 0]^\top$ and  \eqref{eq:comd_zmp} into \eqref{eq:mc_zmp_projection},
the  CoM velocity at the boundary of $\comdArea$ fulfill:
\quickEq{eq:comd_equality}{
  \comdPlane = \pendulumC( \frac{\height{\zmp} - \height{\com}}{\mass g}\inertialForce).
}

% Hence, we summarize the set of balancable CoM velocities:
% \quickEq{eq:comdarea-high}{
% \comdArea \setDef{
%   \comdxy \in \RRv{2}
% }{
%   \text{ $\comdxy$ within the boundary defined by \eqref{eq:comd_equality}}
% }.
% }
Note that \eqref{eq:comd_equality} is $6\numContacts$-dimensional represented, as the resultant force $\inertialForce$ is represented by the sustained contacts' wrenches through \eqref{eq:sum_wrench}.

\subsection{Two-dimensional space representation}
\label{sec:lp_projection}

\begin{figure}[htbp!]
  \vspace{-3mm}
  \centering
  \includegraphics[width=0.98\columnwidth]{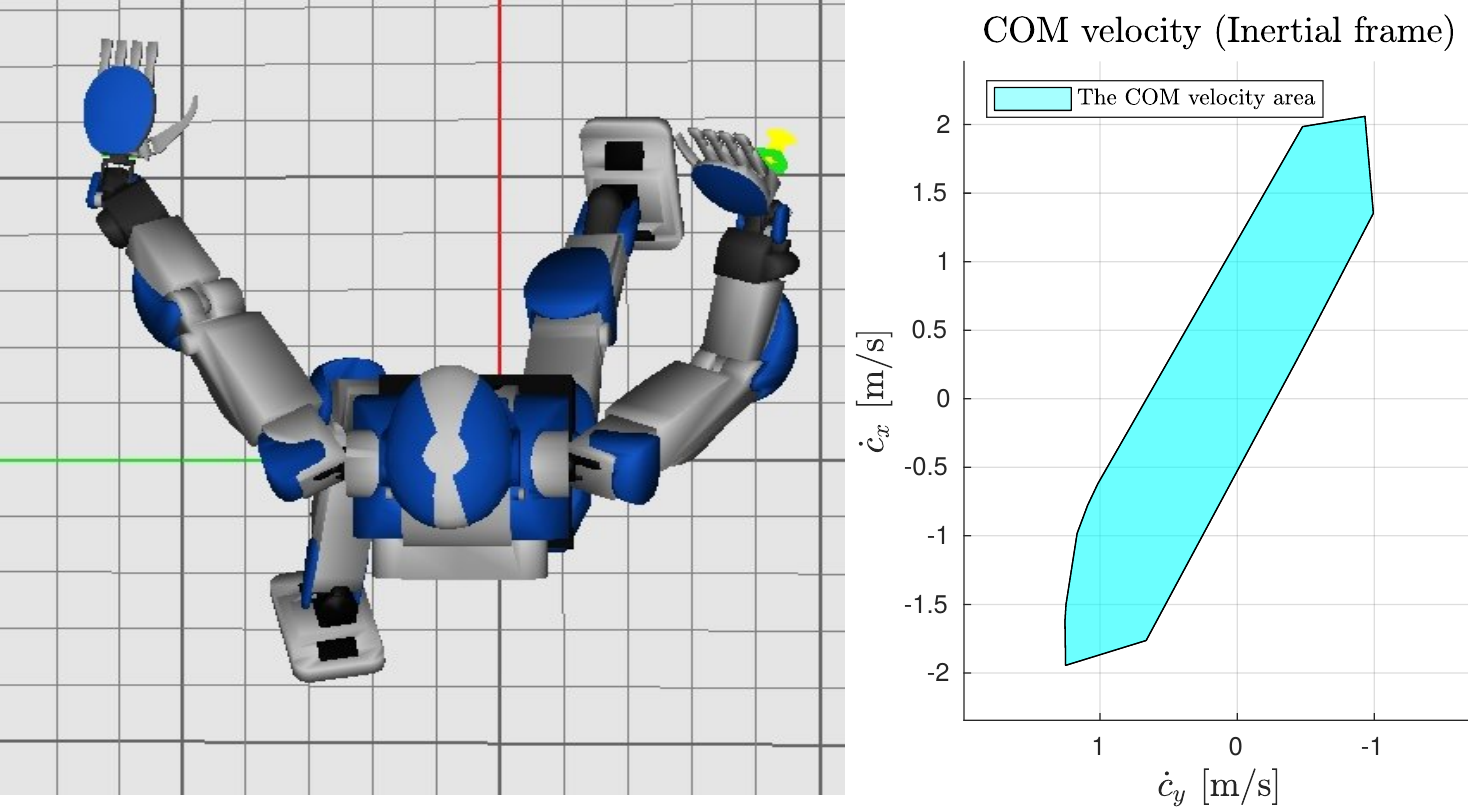}
  \caption{
    The cyan area represents the CoM velocity area $\comdArea$, obtained by iteratively solving (\ref{eq:zmp-projection-obj}-\ref{eq:zmp-projection-def}) for the joint and contact configurations shown on the left (the same stance as in Fig.~\ref{fig:sample_posture}).
  }
  \label{fig:comd-area-example}
\end{figure}

The linear constraints (\ref{eq:cwc_def}, \ref{eq:torque_limits}, \ref{eq:angular_momentum_assumption_constraint}, \ref{eq:comd_equality}) represent the balancable CoM velocities $\comdArea$ in the contact wrench space with a dimension of  $6\numContacts$.
In order to constrain the two-dimensional constraint
$\picomdxy \in \comdArea$,
we  project  $\comdArea$ on the tangent plane of the inertial frame following the ray-shooting algorithm in~\cite{bretl2008tro}.
Namely, to obtain each vertex of $\comdArea$, we iteratively solve the following LP:
\begin{align}
  \label{eq:zmp-projection-obj}
  \max_{\wrench, \bs{y}} \quad \bs{u}_{\theta}^\top \bs{y}& 
  \\
  \label{eq:zmp-projection-inequality}
  \mbox{s.t.}\quad
  A \wrench &\leq B\\ 
  \label{eq:zmp-projection-equality}
  C \wrench  &= d\\
  \label{eq:zmp-projection-def}
  \bs{y} &=  E\wrench + f,
\end{align}
where the optimization variable
$\bs{y}\in \RRv{2}$  denotes a vertex of $\comdArea$
along a given ray (direction) specified by the unit vector  $\bs{u}_{\theta} \in \RRv{2}$. As an example, \figRef{fig:comd-area-example} displays the $\comdArea$ for the HRP-4 stance in \figRef{fig:sample_posture}.

The inequalities \eqref{eq:zmp-projection-inequality} include the joint torque limits \eqref{eq:torque_limits} and collect the CWC constraint \eqref{eq:cwc_def} for  the    $\numContacts$ sustained contacts.
The equality \eqref{eq:zmp-projection-equality}  reformulates \eqref{eq:angular_momentum_assumption_constraint}:
$$
\resizebox{.98\hsize}{!}{$
\underbrace{
\frac{1}{\mass \cdot \gValue} 
\begin{bmatrix} 
\cross{\nz}{} & \height{\com} \cdot \identityMatrix  \\
\nz^\top & -(\nz \times \bs{c})^\top
\end{bmatrix} \cdot \idMatrix{\wrench} 
}_{C}
\underbrace{\begin{bmatrix}
\wrench_1 \\ 
\vdots\\ 
\wrench_{\numContacts}
\end{bmatrix}}_{\wrench}
 = 
 \underbrace{
   \begin{bmatrix}
\mass g\com  %% + \cross{\nz}{{\angularMomentumD}_{\com}}
\\
0 %% \innerP{\nz}{{\angularMomentumD}_{\com}}
\end{bmatrix}
}_{d}.
$}
$$
Similarly, the equality \eqref{eq:zmp-projection-def} reformulates \eqref{eq:comd_equality}:
$$
\vectorTwo{\comd_x}{\comd_y}
 = \underbrace{\frac{\pendulumC(h -\height{\com}) }{\mass \cdot g}\idMatrix{\wrench, [:2,:]}}_{E} 
\underbrace{\begin{bmatrix}
\wrench_1 \\ 
\vdots\\ 
\wrench_{\numContacts}
\end{bmatrix}}_{\wrench}  +
\underbrace{
  \pendulumC\com
}_{f},
$$
where $h$ denotes the height of the projection plane, i.e., $h=0$; see \remarkRef{remark:zmp_height}.
As the resultant force corresponds to the  first two rows of $\inertialWrench$,  $\idMatrix{\wrench, [:2,:]}$ denotes the corresponding part of  $\idMatrix{\wrench}$.

% In addition to fulfilling \eqref{eq:cwc_def} for the contacts, the rest of the inequalities \eqref{eq:zmp-projection-inequality} restrict the contact wrenches according to the robot equations of motion, see \secRef{sec:eqm_constraint}.

\begin{remark} \label{remark:zmp_height}
We are free to choose the height  $h$ of the projection plane as long as $h \neq \height{\com}$. In this paper, we apply $h=0$, which is lower than the CoM height $\height{\com}$. Hence, the inverted pendulum dynamics $\comd = - \pendulumC (\com - \dcm)$ is appropriate. Otherwise, we  need the pendulum dynamics: $\comd =  \pendulumC (\com - \dcm)$ due to the flipped sign of $ h - \height{\com}$; see the details in~\cite{caron2016tro}.    \backfill
\end{remark}

%% \begin{remark} \label{remark:not_determined}
%%   The tangential post-impact CoM velocity varies concerning the impact configurations, e.g., contact velocity see \figRef{fig:gradient_y}, and the impact friction coefficient see  the attached video.
%%   Thus, we can not fix the post-impact CoM velocity normal and
%%   take its intersection with  $\zmpArea$ to find
%%   the ZMP bounds  $\doubleBound{\zmp}$ for (\ref{eq:def_controllable_region_bounds}-\ref{eq:def_stable_standing_region_bounds_special_case}).    \backfill
%% \end{remark}

%% $
%% \comdArea
%% $

\section{Solving the contact velocity}
\label{sec:contactvel}
This section introduces the optimization problem for solving the optimal contact velocity.
\secRef{sec:impact-model} predicts the set of post-impact CoM velocities $\picomdSet$ according to the frictional impact mechanics in three dimensions, and \secRef{sec:control-formulation} formulates the inequalities that can impose $\picomdSet \subseteq \comdArea$ after the impact. \secRef{sec:qp} summarizes the detailed steps.

\subsection{The whole-body impact model}
\label{sec:impact-model}
The  recent  analytical computation \cite{wang2020ijrr} of the set of candidate impulses $\set{\impulse}$ base on the following assumptions:
\begin{enumerate}
  % \item A sustained contact has a Each contact surface is planar.
\item \label{assump:force} The impact induces significant impulsive contact forces and negligible contact torques \cite{stronge2000book}\cite{chatterjee1998jam}.
  Recently, Gong and Grizzle~\cite{gong2020angular} validated this assumption with the biped Cassie robot.
\item The impulse fulfills Coulomb's friction law~\cite{jia2019ijrr,stronge2000book}.

\item The impacting end-effector has a tiny contact  area compared to the robot dimensions such that a point contact model is appropriate \cite{chatterjee1998jam}.
\item The impact detection is timely such that the impacting end-effector can immediately pull back without exerting additional impulse.
%% \item As soon as the impact is detected, the robot immediately pulls back the impacting end-effector without applying sustained forces.
\item \label{assump:crb}
  The robot is kinematically controlled with high gains, enabling it to behave like a composite-rigid body (CRB) during an impact event.
  This assumption applies to robots that are actuated with electrical motors and high reduction-ratio gearboxes \cite{wang2022icra,wang2022ral}.
\end{enumerate}
%% Depending on the robot joint configuration and the contact velocity,
The intersection between the friction cone and the planes of restitution (due to the uncertain restitution coefficient $\coefR \in [\lowerBound{\coefR}, \upperBound{\coefR}]$)
formulates the impulse set $\set{\impulse}$ \cite[Sec.~4.1]{wang2020ijrr}.
% The CoM velocity area $\comdArea$ locates at the inertial frame

According to assumption~\ref{assump:crb}), during the impact event the robot behaves like a rigid body. Thus, the 6-dimensional velocity transform from
the contact point frame $\cframe{\contactPoint}$ to the CoM writes:
$$
\twistTransform{\contactPoint}{\com} = \twistTransformDef{\contactPoint}{\com}.
$$
According to assumption~\ref{assump:force}), the impact does not exert sudden change of momentum.
Thus,  we can compute the set of candidate CoM velocity jumps $\set{\jump \comdxy}$ utilizing the upper-right corner of $\twistTransform{\contactPoint}{\com}$  as:
\quickEq{eq:comd_set}{
  \set{\jump \comd} \setDef{
    \jump \comd \in \RRv{3}
  }{
    \begin{bmatrix}
      \jump \comd % = \jump \bodyTV{\inertialFrame}{\com}
      = \frac{1}{\mass} \rotation{\com}{\contactPoint}\impulse,\\
      \impulse \in \set{\impulse}.   
    \end{bmatrix}
  },
}
where $\mass \in \RRv{+}$ denotes the robot mass, $\rotation{\com}{\contactPoint}$ denotes the rotation from CoM to the contact point, and the impulse set $\set{\impulse}$ is taken from \cite[Sec.~4.1]{wang2020ijrr}.

% As the LIPM model assumes fixed CoM height
% and we project the CoM velocity on the ground $\com_z = 0$, 
% we intersect $\set{\jump \comd}$ with the ground plane to define the candidate planar CoM velocity jumps:
% \quickEq{eq:comd_jump_set}{
%   \set{\jump \comdxy} \setDef{
%     \jump \comdxy \in \RRv{2}
%   }{
%     \begin{bmatrix}
%       \innerP{\basisVec{z}}{\jump\comd}
%       = 0\\
%       \jump \comd \in \set{\jump \comd}   
%     \end{bmatrix}
%   }.
% }
Knowing the pre-impact CoM velocity $\preImpact{\comd}_{xy}$, the set of post-impact  planar CoM velocities are computed as:
\quickEq{eq:post_comd_set}{
  \picomdSet \setDef{
    \picomdxy \in \RRv{2}
  }{
    \begin{bmatrix}
      \picomdxy = \preImpact{\comd}_{xy} + \jump \comdxy
      \\
      \jump \comdxy \in \set{\jump \comdxy} 
    \end{bmatrix}
  }.
}

%% The task-space velocity-to-impulse mapping
%% \paragraph{Impacting object with finite mass}
%% \todo{
%%   When a floating-base robot hits an object with finite mass. 
%% }

\subsection{The post-impact CoM velocity constraint}
\label{sec:control-formulation}
%% \todo{We might have to elaborate more on this point to highlight our own contribution.}
According to the phase-plane analysis \cite{sugihara2009icra,stephens2011thesis},
as long as the post-impact CoM velocity $\picomdxy$ is within the CoM velocity area $\comdArea$:
$$
  \picomdxy \in \comdArea,
$$
the robot can regulate CoM velocity to the origin by the sustained contact wrenches.

Since $\picomdSet$ is a convex set, see \remarkRef{remark:convexity}, as long as each vertex of  $\picomdSet$  is within $\comdArea$,  the entire set  $\picomdSet$ fulfills 
\quickEq{eq:impact-aware-comd}{
  \picomdSet \subset \comdArea.
}
%% Given an on-purpose  impact
%% , e.g., tracking a high contact velocity reference $\reff{\contactVel} \in \RRv{3}$ in task space, 
%% if a whole-body QP controller can regulate the contact velocity $\jacobian\optimal{\jvelocities}$ while fulfilling  the constraint \eqref{eq:impact-aware-comd}, the impact will not lead to a fall.   
In the following, we show how to impose the constraint \eqref{eq:impact-aware-comd} in an optimization problem. 
%% employ vertices of the impulse set $\set{\impulse}$ as optimization variables, and
\begin{remark}
  \label{remark:convexity}
  $\picomdSet$ is convex due to: (1) the impulse set $\set{\impulse}$ is convex by construction. (2) the affine operations, translations and rotations in \eqref{eq:comd_set} and \eqref{eq:post_comd_set} preserves convexity.
  \backfill
\end{remark}

\begin{figure}[htbp!]
  \centering
  \includegraphics[width=0.5\columnwidth]{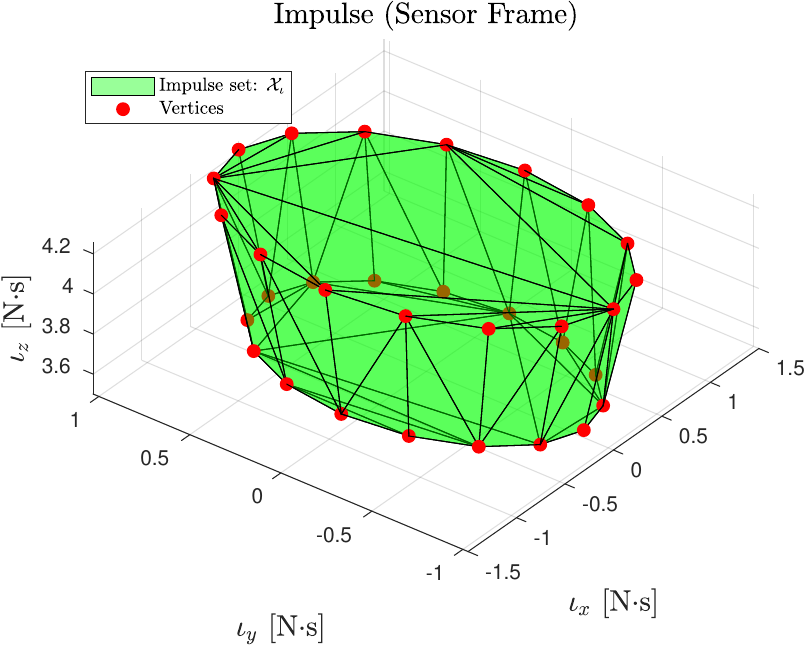}
  \includegraphics[width=0.4\columnwidth]{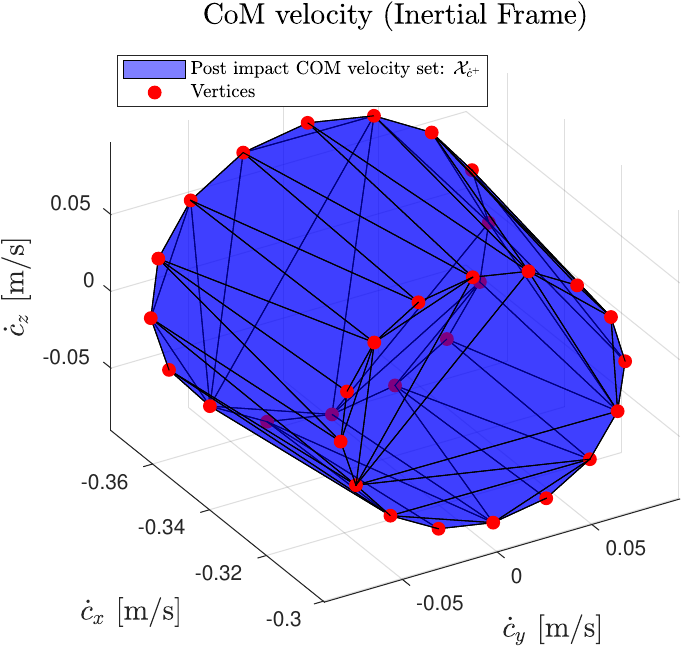}
  \caption{
    For the HRP-4 robot given in Fig.~\ref{fig:sample_posture}, we choose $\fConeNum$ = 16 and visualize  $2\fConeNum$ vertices (red dots) of impulse set $\set{\impulse}$ and the CoM velocity set $\set{\comd}$. The friction coefficient at the impact surface is $0.24$, and the restitution coefficients are limited to $\coefR \in [0, 0.2]$.
}
\label{fig:impulse_set}
\end{figure}

We discritize the Coulumb's friction cone with $\fConeNum$ vertices.
The impulse set $\set{\impulse}$ consists of the interior of the intersection between
the two planes of restitution \cite[Sec. 4.1.3]{wang2020ijrr} and the Coulumb's friction cone, e.g., \figRef{fig:impulse_set}.
Thus, $\set{\impulse}$ has $2\fConeNum$ vertices:
\quickEq{eq:vertices}{
\begin{cases}
  \impulse_{2i} &= \cone{(:,i)}\generatorScalar{2i}  \\
  \impulse_{2i + 1} &= \cone{(:,i)}\generatorScalar{2i + 1} \\
\end{cases} \text{for}~i = 0, \ldots, \fConeNum -1,
}
where $\cone{(:,i)} \in \RRv{3}$ denotes the $i$th column of the Coulumb's friction cone $\cone{\coefF} \in \RRm{3}{\fConeNum}$, and  $\generatorScalar{2i}, \generatorScalar{2i + 1} \in \RRv{+}$ denote the positive scalars. The vertices $\impulse_{i}$ for $i = 0,\ldots, 2\fConeNum - 1$  also fulfill the planes of restitution:
\quickEq{eq:por}{
\begin{aligned}
  \upperBound{\coefR} \postImpact{\contactVel}_z  &= - \preImpact{\contactVel_z} + \transpose{\basisVec{z}} \iim \impulse_{2i}, \\
       \lowerBound{{\coefR}} \postImpact{\contactVel}_z &= - \preImpact{\contactVel_z} + \transpose{\basisVec{z}} \iim \impulse_{2i + 1},
\end{aligned} \text{for } i = 0,\ldots, \fConeNum - 1.
}
%%  have to fulfill the following equalities according to \eqref{eq:final_impulse_set}: 
where the inverse inertia $\iim \in \RRm{3}{3}$ denotes the impulse-to-velocity mapping and keeps constant as the robot configuration does not change during the impact event, $\preImpact{\contactVel_z}$ denotes the pre-impact contact velocity along the normal direction  $\basisVec{z} \in \RRv{3,}$\footnote{According to impact mechanics\cite{stronge2000book,wang2022ral}, the contact frame's Z-axis aligns with the impact's normal direction.}, and the uncertain restitution coefficient fulfills $\coefR \in [\lowerBound{\coefR}, \upperBound{\coefR}]$.
% \begin{remark}
%   We can not analytically project $\picomdSet$ on the X-Y plane. Thus, we simply restrict the entire set.
%   \backfill
% \end{remark}

Therefore, employing 
$\generatorScalar{i}$, for $i=0,\ldots,2(\fConeNum -1)$
as optimization variables,
the solver is aware of the set of post-impact impulses $\set{\impulse}$ according to the \highlightblue{\text{impulse-set constraint}}:
% \eqref{eq:impulse-set-constraints} modify the solver's search space such that \eqref{eq:impact-aware-comd} is always respected.
\begin{equation}
  \label{eq:impulse-set-constraints}
  \begin{aligned}
    & \highlight{\text{For }i = 0, \ldots, \fConeNum -1:} \\ 
    & \highlight{\text{Friction cone:}}
  \begin{cases}
    \generator{2i} &\geq \zeroVector \\
    \generator{2i + 1} &\geq \zeroVector,
  \end{cases} \\
  & \highlight{\text{Plane of restitution: \eqref{eq:por}}}, \\
  & (\upperBound{\coefR} + 1)  J_3 \jvelocities = \transpose{\basisVec{z}} \iim \cone{(:,i)}\generator{2i},
    \\
  &(\lowerBound{{\coefR}} + 1) J_3 \jvelocities = \transpose{\basisVec{z}} \iim \cone{(:,i)}\generator{2i+1}, 
  \end{aligned}
\end{equation}
where $J_3 \jvelocities$ computes the contact velocity along the normal direction.

We denote the $2i$-th vertex of $\picomdSet$ as $\picomdVertex{2i} \in \RRv{2}$. 
Given the mappings \eqref{eq:comd_set}, \eqref{eq:post_comd_set}, and
the impulse vertices $\impulse_{2i}, \impulse_{2i+1}$ defined by \eqref{eq:vertices} and \eqref{eq:por}, $\picomdVertex{2i}$ writes:
$$
\begin{aligned}
\picomdVertex{2i} &= \preImpact{\comVelPlane} +  \frac{1}{\mass}
  \begin{bmatrix}
    1,& 0,& 0\\
    0,& 1,& 0
  \end{bmatrix}
  \rotation{\com}{\contactPoint}\cone{(:,i)}\generator{2i},\\
  \picomdVertex{2i+1} &= \preImpact{\comVelPlane} +  \frac{1}{\mass}
  \begin{bmatrix}
    1,& 0,& 0\\
    0,& 1,& 0
  \end{bmatrix}
  \rotation{\com}{\contactPoint}\cone{(:,i)}\generator{2i+1}.
\end{aligned}
$$
Therefore, we reformulate \eqref{eq:impact-aware-comd} as the following \highlightblue{\text{post-impact  }}  \highlightblue{\text{ CoM velocity constraint}} for $i = 0, \ldots, \fConeNum -1$:
\quickEq{eq:picomd-constraint-full}{
  \begin{aligned}
    & \underbrace{\preImpact{\comVelPlane} +  \frac{1}{\mass}
  \begin{bmatrix}
    1,& 0,& 0\\
    0,& 1,& 0
  \end{bmatrix}
  \rotation{\com}{\contactPoint}\cone{(:,i)}\generator{2i}}_{\picomdVertex{2i}}   \in \comdArea;\\
  & \underbrace{\preImpact{\comVelPlane} +  \frac{1}{\mass}
  \begin{bmatrix}
    1,& 0,& 0\\
    0,& 1,& 0
  \end{bmatrix}
  \rotation{\com}{\contactPoint}\cone{(:,i)}\generator{2i + 1}}_{\picomdVertex{2i+1}}   \in \comdArea,\\
  \end{aligned},
}
which modifies the solver's search space to impose the balance condition \eqref{eq:impact-aware-comd}.
We assume  $\comdArea$ has $n_c$ vertices. In order to implement  $\picomdVertex{i} \in \comdArea$, we have to re-write the vertex-represented $\comdArea$ as the half-space representation $[G_{\comdArea} \in \RRm{n_c}{2}, h_{\comdArea} \in \RRv{n_c}]$, and reformulate \eqref{eq:picomd-constraint-full} as:
\quickEq{eq:picomd-constraint}{
  \begin{aligned}
  G_{\comdArea} \picomdVertex{i} \leq h_{\comdArea}, 
  \end{aligned} \quad \text{for} \quad i=1,\ldots, 2\fConeNum -1.
}
There are standard approaches to convert between vertex representation and the half-space representation of a polytope, e.g., the double description method \cite{fukuda2005double}.

\subsection{Quadratic program formulation}
\label{sec:qp}

As a minimalistic example, we can formulate an optimization problem which takes the following form:
  \quickEq{qp:velocity}{
    \begin{aligned}
      \min_{\jvelocities, \generator{\coefF}} \quad & \twoNorm{\jacobian\jvelocities  - \reff{\contactVel}}
      \\
      \mbox{s.t.} \quad
      &\highlightblue{\text{The impulse-set constraint \eqref{eq:impulse-set-constraints}},}\\
      &\highlightblue{\text{The post-impact CoM velocity constraint \eqref{eq:picomd-constraint}},}\\
      \quad& \text{Other impact-aware constraints, see Remark~\ref{remark:impulse-torque}}.
    \end{aligned}
  }
  where the high-lighted constraints are mandatory to meet the proposed balance criteria. See the detailed formulation steps in~\algRef{algorithm:qp}. 

\begin{figure}[htbp!]
\vspace{-2mm}
  \centering
  \includegraphics[width=1.0\columnwidth]{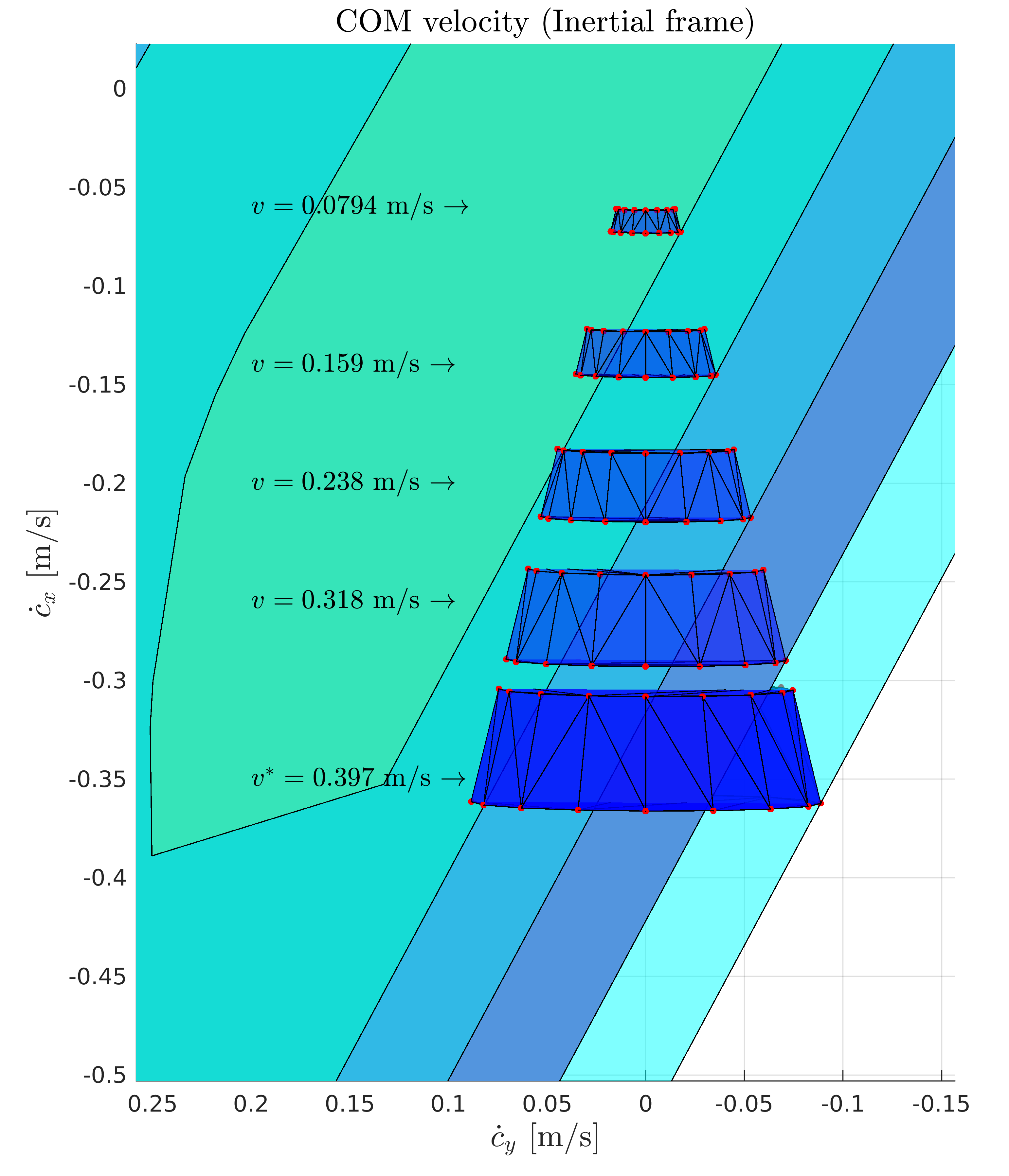}
  \caption{
    Illustration of the relation between contact velocity $\jacobian\jvelocities$ and the corresponding post-impact CoM velocity set $\picomdSet$. As  $\jacobian\jvelocities$ increases, the solver identified the maximum contact velocity $\jacobian\jvelocities = 0.397$\unitVelTS without violating the condition $\picomdSet \subset \comdArea$.
    The different colored areas indicate corresponding hypothetical CoM velocity areas that would be defined, if the contact velocities $0.0794, 0.159, 0.238, 0.318$\unitVelTS were found to be optimal.
}
\label{fig:solving-opt}
\vspace{-3mm}
\end{figure}

  By solving~\eqref{qp:velocity}, we can obtain the maximum feasible contact velocity $\jacobian\optimal{\jvelocities}$ for a whole-body controller~\cite{koolen2016ijhr,bouyarmane2018tac,bouyarmane2019tro}. 
  As an example,
  formulating and solving \eqref{qp:velocity} leads to the contact velocity $\jacobian \optimal{\jvelocities} = 0.397$\unitVelTS for  the HRP-4 stance in \figRef{fig:sample_posture}.
  Applying this velocity  would result in a vertex of the set $\picomdSet$ (visualized in \figRef{fig:impulse_set}) overlapping with the boundary of  $\comdArea$ in \figRef{fig:solving-opt}, thereby satisfying the condition  \eqref{eq:impact-aware-comd}.
  
  To illustrate the solution's optimality, we additionally plotted the sets $\picomdSet$ resulted from a series of increasing  contact velocities $0.0794, 0.159, 0.238, 0.318$\unitVelTS in \figRef{fig:solving-opt}, assuming that the background colored areas are the actual CoM velocity area $\comdArea$.
  Hence, the optimizer would increase the contact velocity to $0.397$\unitVelTS.

\begin{figure}[htbp!]
  \centering
  \includegraphics[width=1.0\columnwidth]{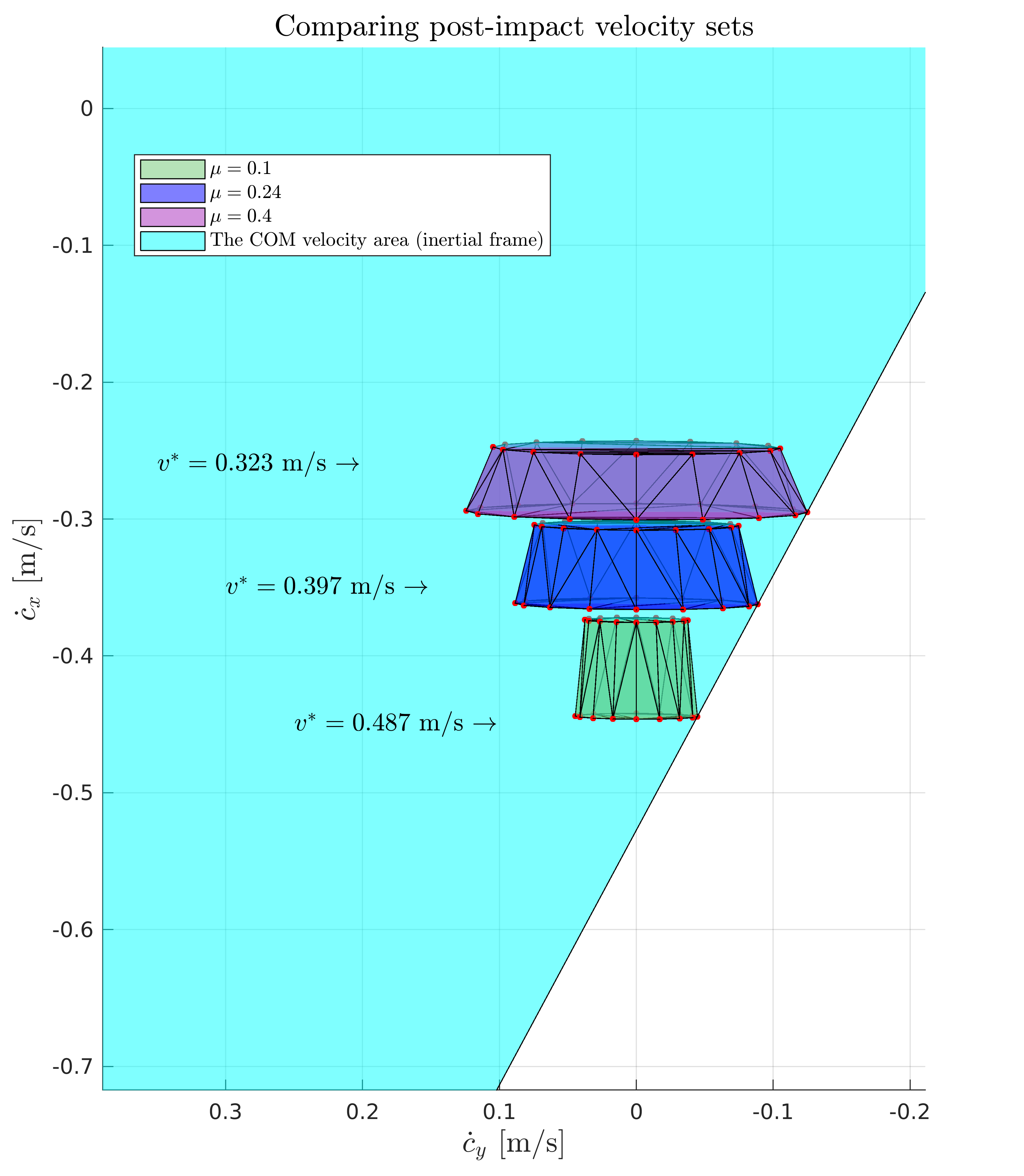}
  \caption{
    This figure compares the optimal solutions for friction coefficients of $0.1$, $0.24$, and $0.4$. As the geometric size of the post-impact CoM velocity set $\picomdSet$ varies with the friction coefficient, the solver identifies different optimal contact velocities.
    %Comparison of different optimal solutions for friction coefficients of $0.1$, $0.24$, and $0.4$. The solver identifies different optimal contact velocities due to the varying geometric sizes of the post-impact CoM velocity set $\picomdSet$.
}
\label{fig:miu-compare}
\vspace{-3mm}
\end{figure}

  The geometric size of the set $\picomdSet$ varies with respect to the friction coefficient at the impact surface. We show different optimal contact velocities $\jacobian \optimal{\jvelocities}$ resulting from the variation in friction coefficient \figRef{fig:miu-compare}.

\begin{remark}
  \label{remark:impulse-torque}
To protect the hardware resilience bounds, we can additionally apply the impact-aware constraints for $i = 0,\ldots, \fConeNum - 1$:
\begin{equation}
  \label{eq:additional-impulse-constraints}
  \begin{aligned}
    & \highlight{\text{Post-impact joint velocity: \cite[eq.~29]{wang2020ijrr}}}, \\
    &\lowerBound{\jvelocities} \leq \preImpact{\jvelocities} + \pseudoInverseRowDef{\jacobian} \iim \cone{(:,i)}\generator{2i} 
    \leq \upperBound{\jvelocities},\\
    & \lowerBound{\jvelocities} \leq \preImpact{\jvelocities} + \pseudoInverseRowDef{\jacobian} \iim \cone{(:,i)}\generator{2i+1} 
    \leq \upperBound{\jvelocities},  
    \\
    & \highlight{\text{Post-impact joint torque: \cite[eq.~30]{wang2020ijrr}}}, \\
    & \lowerBound{\torque} \leq \preImpact{\torque} +  \frac{\coefPF}{\impactDuration}  \transpose{\jacobian}\cone{(:,i)}\generator{2i}    \leq \upperBound{\torque}, \\
    & \lowerBound{\torque} \leq \preImpact{\torque} +  \frac{\coefPF}{\impactDuration}  \transpose{\jacobian}\cone{(:,i)}\generator{2i+1}    \leq \upperBound{\torque} \\ 
  \end{aligned},
\end{equation}
where the impact duration $\impactDuration$ and the positive scalar $\coefPF$ facilitate the prediction of impulsive torque \cite[Sec.~4.3]{wang2020ijrr}. \backfill
\end{remark}

\begin{algorithm}
  \caption{Solve the maximum balance-constraint-aware contact velocity $\jacobian\optimal{\jvelocities}$}
  \label{algorithm:qp}
%%   \hspace*{\algorithmicindent}
  \textbf{Inputs:}
  (1) The robot's joint positions $\jangles$, velocities $\jvelocities$; (2) The contact configurations, i.e., the contact area's geometric size and friction coefficients, for computing the contact wrench cone \eqref{eq:cwc_def}; (3) The impact end-effector's friction coefficient; (4) The reference contact velocity $\reff{\contactVel} \in \RRv{3}$.\\
%%   The impact configurations listed by \secRef{sec:impact_config}, and we know $\tp{\contactVel} = 0$ from the numerical integration. \\
%%   \hspace*{\algorithmicindent}
  \textbf{Outputs:}
  The maximum contact velocity $\jacobian\optimal{\jvelocities}$ without leading to a fall.
  \begin{algorithmic}[1]
    \Procedure{Initialization }{High-dimensional represented $\comdArea$}
    \State Contact wrench cones \eqref{eq:cwc_def}.
    \State Limited actuation torques \eqref{eq:torque_limits}.
    \State LIPM assumptions \eqref{eq:angular_momentum_assumption_constraint}.
    \State CoM velocity dependence on ZMP \eqref{eq:comd_equality}.
    \EndProcedure{}
    \Procedure{Projection}{}
    \State Iteravtively solving  (\ref{eq:zmp-projection-obj}-\ref{eq:zmp-projection-def}) to obtain vertices of the two-dimensional  $\comdArea$, \cite{bretl2008tro,roux2021icar}.    
    \EndProcedure{}
    \Procedure{Solve and execute $\jacobian\optimal{\jvelocities}$ }{}
    \State
    Solve QP  \eqref{qp:velocity} with impulse set \eqref{eq:impulse-set-constraints} and post-impact CoM velocity \eqref{eq:picomd-constraint} constraints for $\jacobian\optimal{\jvelocities}$
    \State
    Track $\jacobian\optimal{\jvelocities}$ and fulfill other task-space objectives using a whole-body QP, such as \cite{koolen2016ijhr,bouyarmane2019tro}.
    % Formulate another whole-body QP, e.g., \cite{bouyarmane2019tro}, to track   while fulfilling other task-space objectives.
    \EndProcedure{}
  \end{algorithmic}
\end{algorithm}

\section{Validation}
\label{sec:validation}
Employing the full-size humanoid robot HRP-4 with 34 actuated joints, we validate the proposed approach through the following experiments and simulations:

\begin{experiment}
\item \label{experiment:push-wall} {\bf[Experiment: Impacting a Location-Unknown Wall]}
  To show that the CoM velocity is a less-conservative measure than ZMP, we strictly restrict the ZMP with the support polygon $\zmp \in \supportPolygon$ following the impact-aware QP formulation \cite{wang2020ijrr} and found a contact velocity of $0.11$\unitVelTS. 
  In another trial, we tried a significantly higher contact velocity $0.345$\unitVelTS.
  Despite the violation of the ZMP criteria  $\zmp \in \supportPolygon$, the robot maintained its balance and did not fall.
\item \label{experiment:push-recovery} {\bf[Experiment: Push Recovery with Two Non-coplanar Contacts]}
  To validate the proposed CoM velocity area $\comdArea$ in \secRef{sec:comd_area}, we placed the HRP-4 robot on two  non-coplanar contacts. Pushing the robot from different directions, we observed that the ZMP momentarily jumped outside the multi-contact support area \cite{caron2016tro}, while the robot did not fall.
Throughout the experiment, the proposed CoM velocity area condition $ \picomdxy \in \comdArea$ was always respected.
\end{experiment}

\begin{simulation}

\item \label{simulation:multi-stances} {\bf[Simulation: Maximum Contact Velocities for various stances]}  
  We determined the maximum contact velocity by solving problem \eqref{qp:velocity} for three different scenarios: (1) pushing with two coplanar contacts as shown in \figRef{fig:example_one}; (2) pushing with two non-coplanar contacts; (3) kicking with one foot contact. 
%%     for pushing with non-coplanar contacts, and kicking stances.
    % ; see \figRef{fig:collection} and \tableRef{table:collection}.

%% \item \label{simulation:push-wall} {\bf[Simulation: Validating the Proposed Criteria]}
%%   Apply the same configurations, the COM velocity from E.~\ref{experiment:push-wall} fulfills the $\comdArea$ obtained in S.~\ref{simulation:push-wall}, which explains the stable standing observed in  E.~\ref{experiment:push-wall}.
% \item \label{simulation:max-push-wall} {\bf[Simulation: Exploiting the Maximum Push Velocity]}
%   % Employing the same parameters as S.~\ref{simulation:push-wall},
%   Employing the same impact model configurations with S.~\ref{simulation:push-wall}, we identified the maximum contact velocity
%   is about  $0.42$\unitVelTS for the same stance.
\end{simulation}
\subsection{Experiment~\ref{experiment:push-wall} Impacting a Location-Unknown Wall}

\begin{figure}[!htbp]
  \centering
  \includegraphics[width=0.45\textwidth]{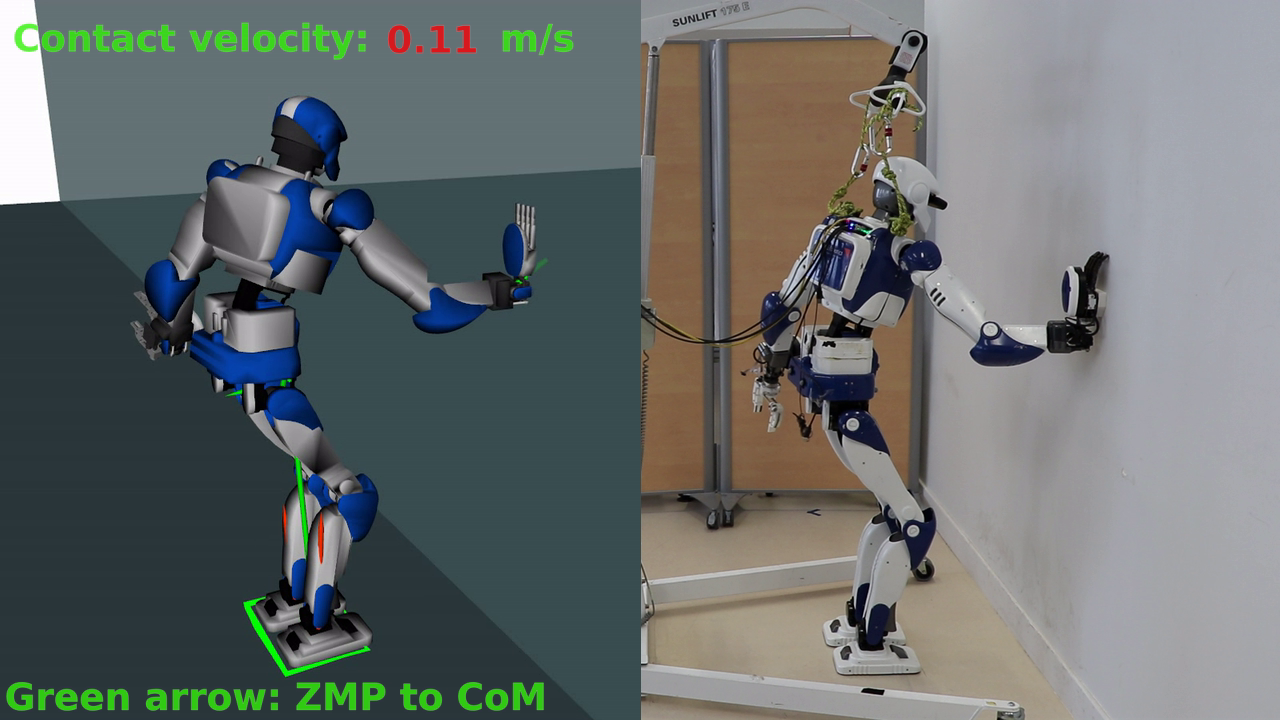}
  \caption{
    The snap shot of the HRP-4 robot hitting the wall at $0.11$\unitVelTS while strictly fulfilling $\zmp \in \supportPolygon$ \url{https://youtu.be/TL34EWORwbU}. The green arrow connects the CoM and the ZMP on the ground. The red arrows indicate the ground reaction forces. 
  }
\label{fig:zmp-push}
\end{figure}

\paragraph{Experiment setup}
The HRP-4 Robot's joints are controlled in position at $1000$\unitHz, while the mid-level QP controller~\cite{bouyarmane2018tac} samples the ATI-45 force-torque sensors mounted on the ankles and palms at  $200$\unitHz. The friction coefficient for the feet contacts is set to  $0.7$, and the robot does not have prior knowledge of the position of the concrete wall shown in~\figRef{fig:zmp-push}.

% . We choose the coefficient of restitution e =
% 0.02, which indicates trivial rebounce and leads to reasonable
% prediction of the impulsive force when the impact duration
% δt is selected as 5 ms.

Optimization-based whole-body controllers rely on closed-form calculations to formulate the optimization problem during each sampling period.
The analytical impact model~\cite{zheng1985jfr} can only predict sudden changes of momentum or impulse $\impulse$.
In order to predict the
sudden change of the ZMP, we
artificially set the impact duration $\impactDuration = 5$\unitMS to
predict the impulsive force as $\jump \force = \frac{\impulse}{\impactDuration}$.

\paragraph{ZMP-based criteria validation}
Our QP controller commanded the HRP-4's right palm to hit a wall with exceptionally-high reference contact velocity $0.8$\unitVelTS,~see~\figRef{fig:zmp-push}. We mounted a 3D printed plastic palm with $3$~cm\xspace thickness.

Employing the predicted  $\jump \force$, the optimization solver reduced the contact velocity to 0.11\unitVelTS, see \figRef{fig:contact-vel-11} to ensure the balance criterion $\zmp \in \supportPolygon$. The robot immediately pulled back the right palm as soon as the force-torque measurements reached 15\unitForce.

\begin{figure}[!htbp]
  \centering
  \includegraphics[width=0.45\textwidth]{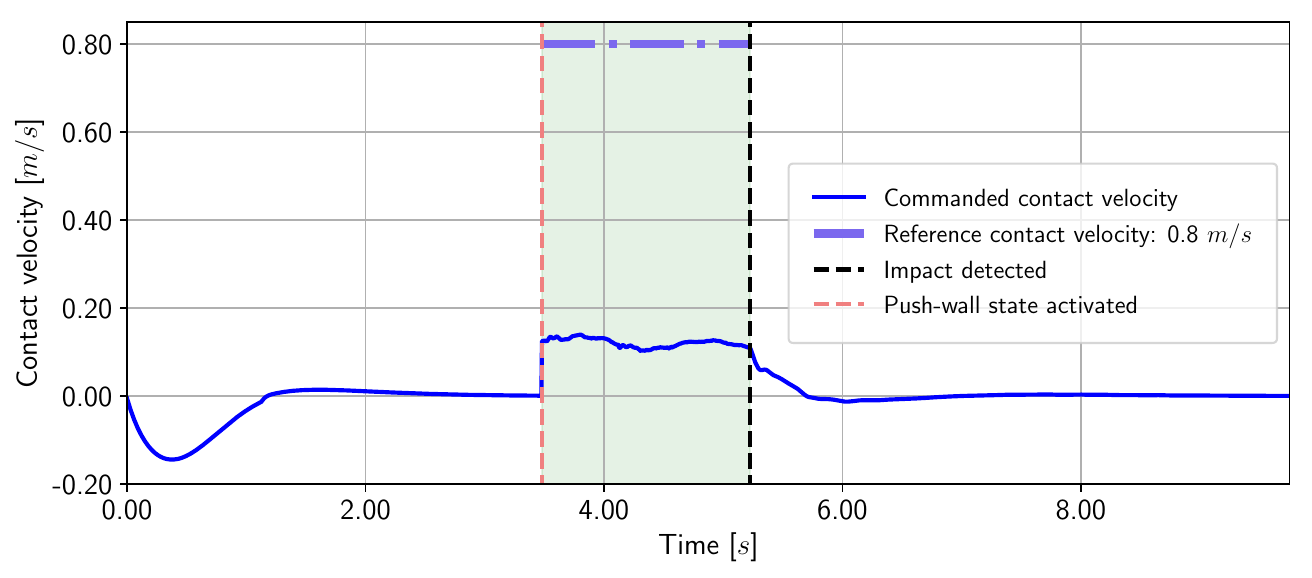}
  \caption{
    In order to meet the ZMP-based balance criterion  $\zmp \in \supportPolygon$,
    the impact-aware QP \cite{wang2020ijrr} reduced the contact velocity to  $0.11$\unitVelTS from a high reference $0.8$\unitVelTS. 
  }
  \label{fig:contact-vel-11}
\end{figure}
\begin{figure}[!htbp]
  \centering
  \includegraphics[width=0.45\textwidth]{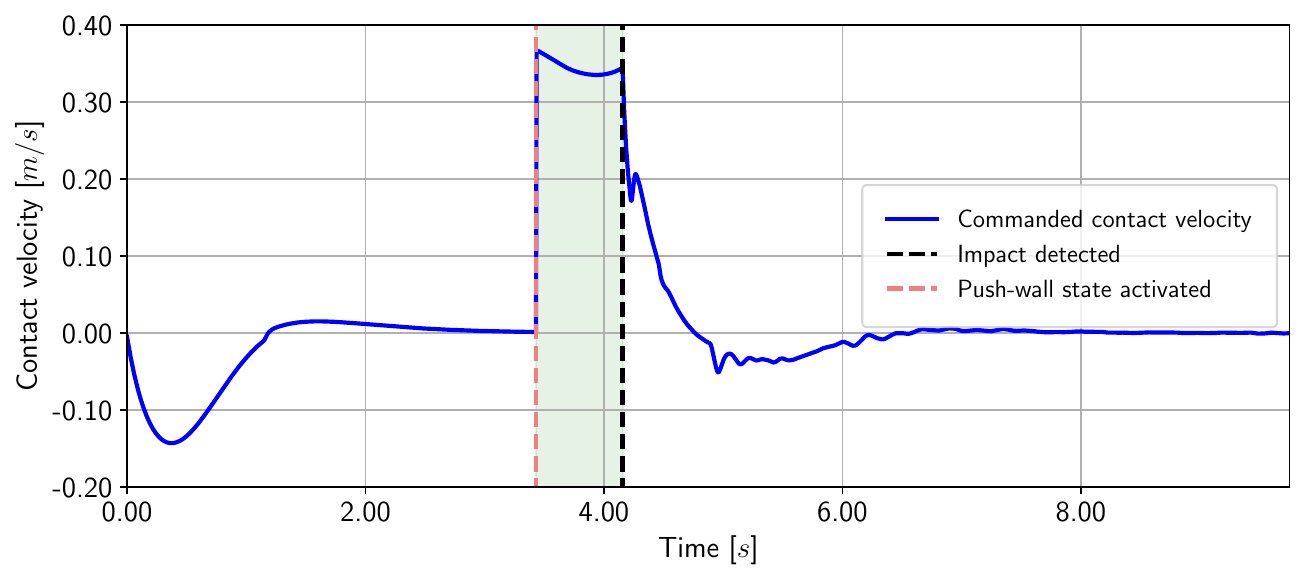}
  \caption{
    Without considering the ZMP-based balance criterion  $\zmp \in \supportPolygon$, we increased the
    contact velocity to  $0.34$\unitVelTS. The robot did not fall as robot regulated the CoM velocity to the origin in the right corner of Fig.~\ref{fig:example_one}.
  }
  \label{fig:contact-vel-11}
\end{figure}

We observed that the  right plam's impact was considerably gentle. Thus, in another trial, we removed the constraint $\zmp \in \supportPolygon$ and changed the contact velocity reference to $0.34$\unitVelTS. Despite the ZMP jumped outside the support polygon, see \figRef{fig:example_one}, the robot did not fall, see the analysis in \secRef{sec:analysis}.

\paragraph{Fulfilling zero-step capture region}
In contrast, according to the CoM velocity, the robot will not fall after the impact at $0.34$\unitVelTS because the CoM velocities strictly lie within the zero-step capture region, as shown in  \figRef{fig:example_one}.

\subsection{Experiment~\ref{experiment:push-recovery} Push recovery on two non-coplanar contacts}

\begin{figure}[!htbp]
  \centering
  \includegraphics[width=0.5\textwidth]{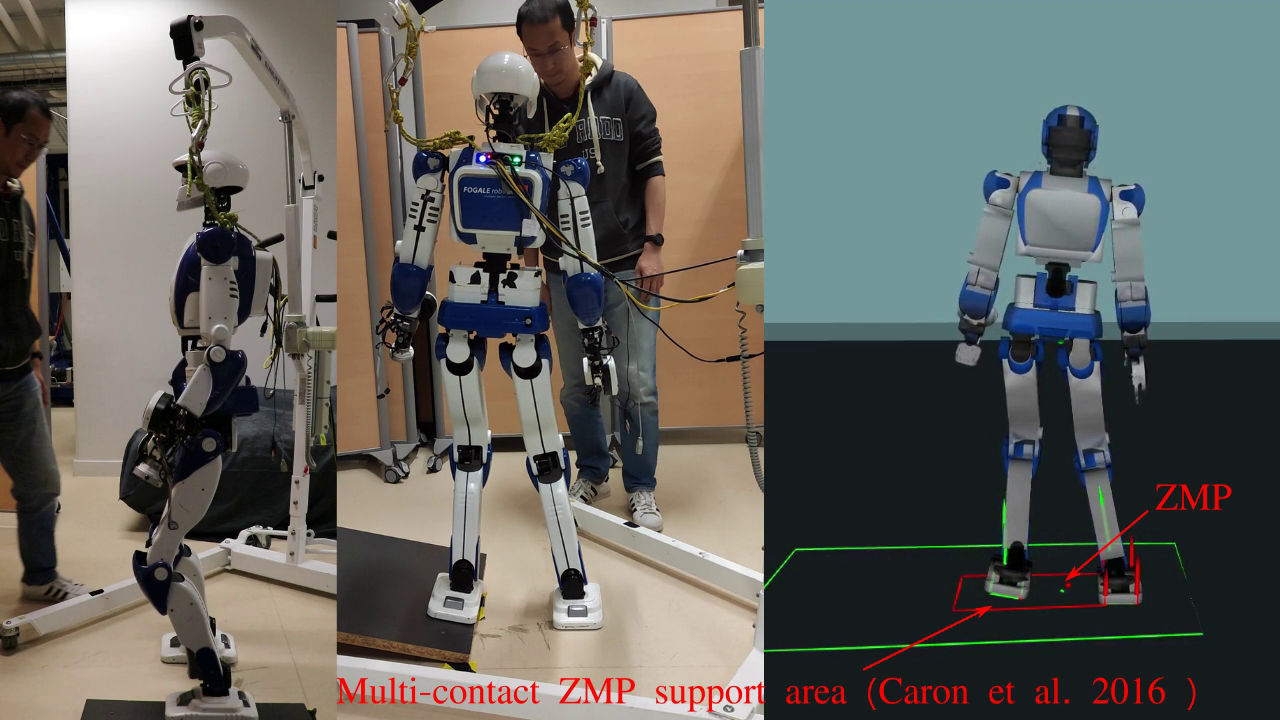}
  \caption{
    From left to right, we show the side, back and simulation view of the HRP-4 robot on two non-coplanar contacts. The red dot and the red rectangle correspond to ZMP $\zmp$ and the ZMP support area $\zmpArea$ by~\cite{caron2016tro}. The green dot and the green rectangle correspond to the CoM velocity $\comdxy$ and the proposed CoM velocity area $\comdArea$. 
  }
\label{fig:non-coplanar-stance}
\end{figure}

\paragraph{Experiment setup}
We placed the HRP-4 robot on two non-coplanar contacts, see \figRef{fig:non-coplanar-stance} and regulated the CoM dynamics with the LIPM stabilizer through the mc$\_$rtc framework\footnote{\url{https://jrl-umi3218.github.io/mc_rtc/tutorials/recipes/lipm-stabilizer.html}}.

\begin{figure}[!htbp]
  \centering
  \includegraphics[width=0.5\textwidth]{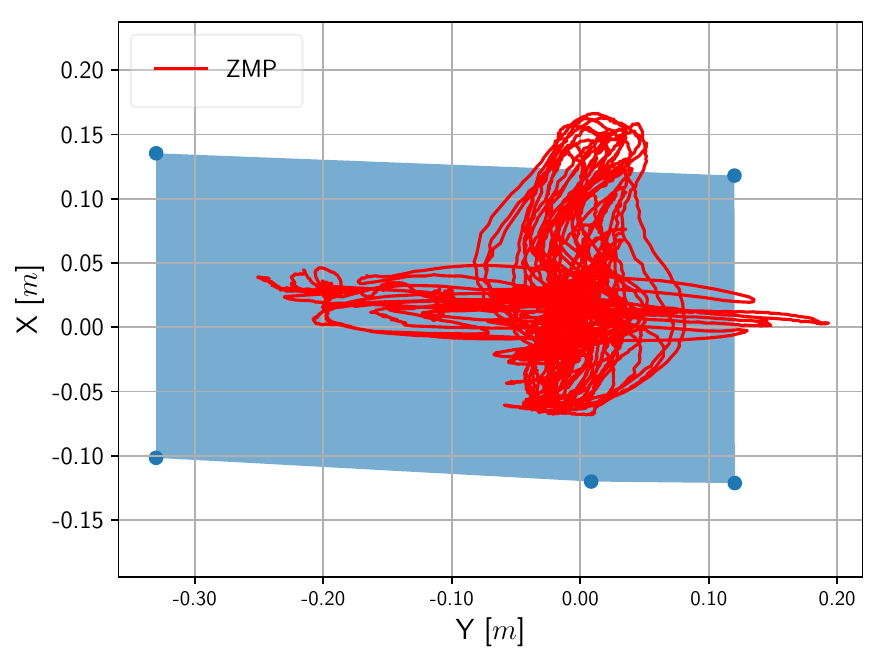}
  \caption{
    The ZMP trajectory (red line) and the ZMP support area   $\zmpArea$ (grey rectangle) during the push-recovery experiment.
    Temporary violations of the ZMP support area, indicated by $\zmp$ values outside of $\zmpArea$, did not result in a fall. 
  }
  \label{fig:zmp-plot}
  \vspace{-3mm}
\end{figure}

The state-of-the-art balance criteria \cite{caron2016tro} for non-copalnar contacts requires ZMP  to stay strictly within the ZMP support area $\zmpArea$, e.g., the light blue region in \figRef{fig:zmp-plot} (or the red polygon in \figRef{fig:non-coplanar-stance}), which is the extension of the support polygon for coplanar contacts. 

\begin{figure}[!htbp]
  \centering
  \includegraphics[width=0.5\textwidth]{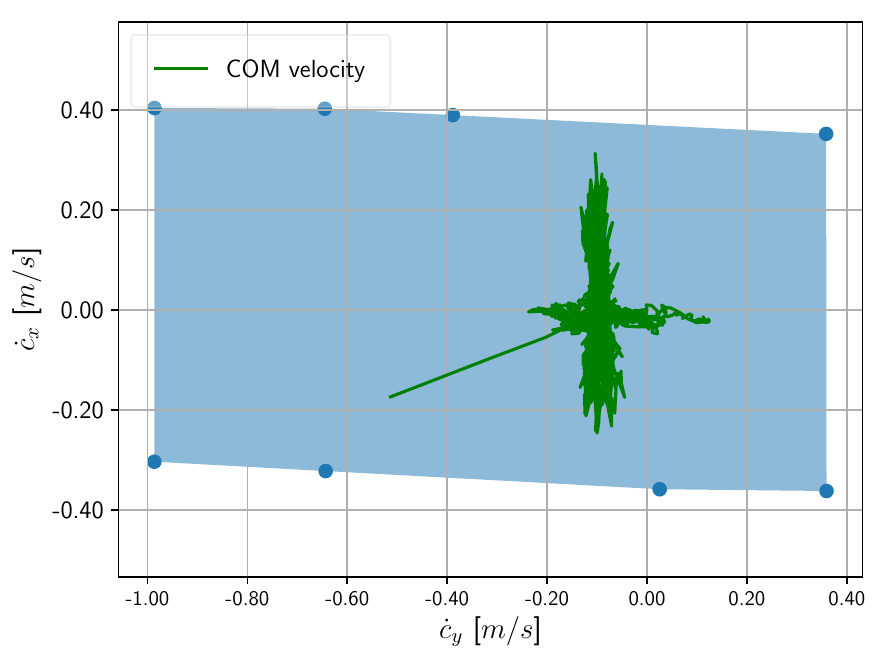}
  \caption{
    Trajectory of the CoM velocity $\comdxy$ plotted against the CoM velocity area $X[\dot{x}_c]$, during the push-recovery experiment. The spike pointing in the down-left direction corresponds to the initial motion, when the robot established non-coplanar contact on the ramp.    
  }
\label{fig:comd-plot}
\end{figure}

\paragraph{Balance criteria comparison}
The operator pushed the HRP-4 robot from front, side, and back, respectively and for multiple times. 
The ZMP temporarily violated the ZMP support area $\zmpArea$ for several times without leading to a fall, see \figRef{fig:zmp-plot}. On the other hand, during the same experiment the CoM velocity  $\comdxy$ strictly fulfilled the CoM velocity area $\comdArea$, see \figRef{fig:comd-plot}. Thus, we conclude that the condition $\picomdxy \in \comdArea$ is more accurate than state-of-the-art ZMP-based balance criteria.


\subsection{Simulation~\ref{simulation:multi-stances} On-purpose impact for various stances}
% \begin{figure}[htbp!]
%   \vspace{-3mm}
%   \centering
%   \includegraphics[width=0.5\columnwidth]{fig/comd-area-pushwall/impulse_set-crop.pdf}
%   \includegraphics[width=0.4\columnwidth]{fig/comd-area-pushwall/COM_Velocity-crop.pdf}
%   \caption{
%     For the HRP-4 robot in Fig.~\ref{fig:example_one}, we visualize vertices (red dots) of impulse set $\set{\impulse}$ and the COM velocity set $\set{\comd}$ when the contact velocity $\jacobian\optimal{\jvelocities}=0.389$\unitVelTS. The friction coefficient at the impact surface is $0.2$, and the restitution coefficients are limited to $\coefR \in [0, 0.2]$.
% } 
% \label{fig:impulse_set_push_wall}
% \end{figure}
% \begin{figure}[htbp!]
%   \vspace{-3mm}
%   \centering
%   % \includegraphics[width=0.8\columnwidth]{fig/comd-area-pushwall/comd-push-wall.pdf}
%   \includegraphics[width=0.8\columnwidth]{fig/comd-area-pushwall/comd-area-comparison.png}
%   \caption{
%     When the  HRP-4 robot in Fig.~\ref{fig:example_one} applies the contact velocity $\jacobian\optimal{\jvelocities}=0.389$\unitVelTS, 
%     vertices of set $\set{\comd}$ overlap the boundaries of $\comdArea$.
%   }
%   \label{fig:comd_push_wall}
% \end{figure}

In order to evaluate the maximum contact velocities for various on-purpose impact tasks, we follow the steps outlined in \algRef{algorithm:qp} to formulate and solve \eqref{qp:velocity}  for the \emph{maximum contact velocity} $\jacobian\optimal{\jvelocities}$  which is the highest contact velocity without breaking \eqref{eq:impact-aware-comd}.

As elaborated in \secRef{sec:qp} and visualized in \figRef{fig:comd-area-example}, applying $\jacobian\optimal{\jvelocities}$  would enable the boundaries of the post-impact CoM velocities $\picomdSet$ and the CoM velocity area $\comdArea$ overlap with each other. 
We will refer to phenomenon as the \emph{overlapping condition} in future discussions.

Additional details are available in the attached video, where  we tuned the joint and contact configurations using a Graphical User Interface and visualize the CoM velocity area $\comdArea$.

% \paragraph{Push-wall with two planar contacts stance}
% One may wonder what the maximum contact velocity would be for the stance shown in \figRef{fig:example_one}.
% Choosing the right palm as the impact end-effector with the friction coefficient $\coefF = 0.2$ and restitution coefficient $\coefR \in [0, 0.2]$, we visualized the set of candidate impulses $\set{\impulse}$  and post-impact COM velocity jumps in \figRef{fig:impulse_set_push_wall}.

% By solving \eqref{qp:velocity} without including the hardware resilience constraint \eqref{eq:additional-impulse-constraints}, we found the contact velocity $\jacobian\optimal{\jvelocities} = 0.389$ \unitVelTS, which enabled the \emph{overlapping condition} in \figRef{fig:comd_push_wall}.

\begin{figure}[htbp!]
  \centering
  \includegraphics[width=0.8\columnwidth]{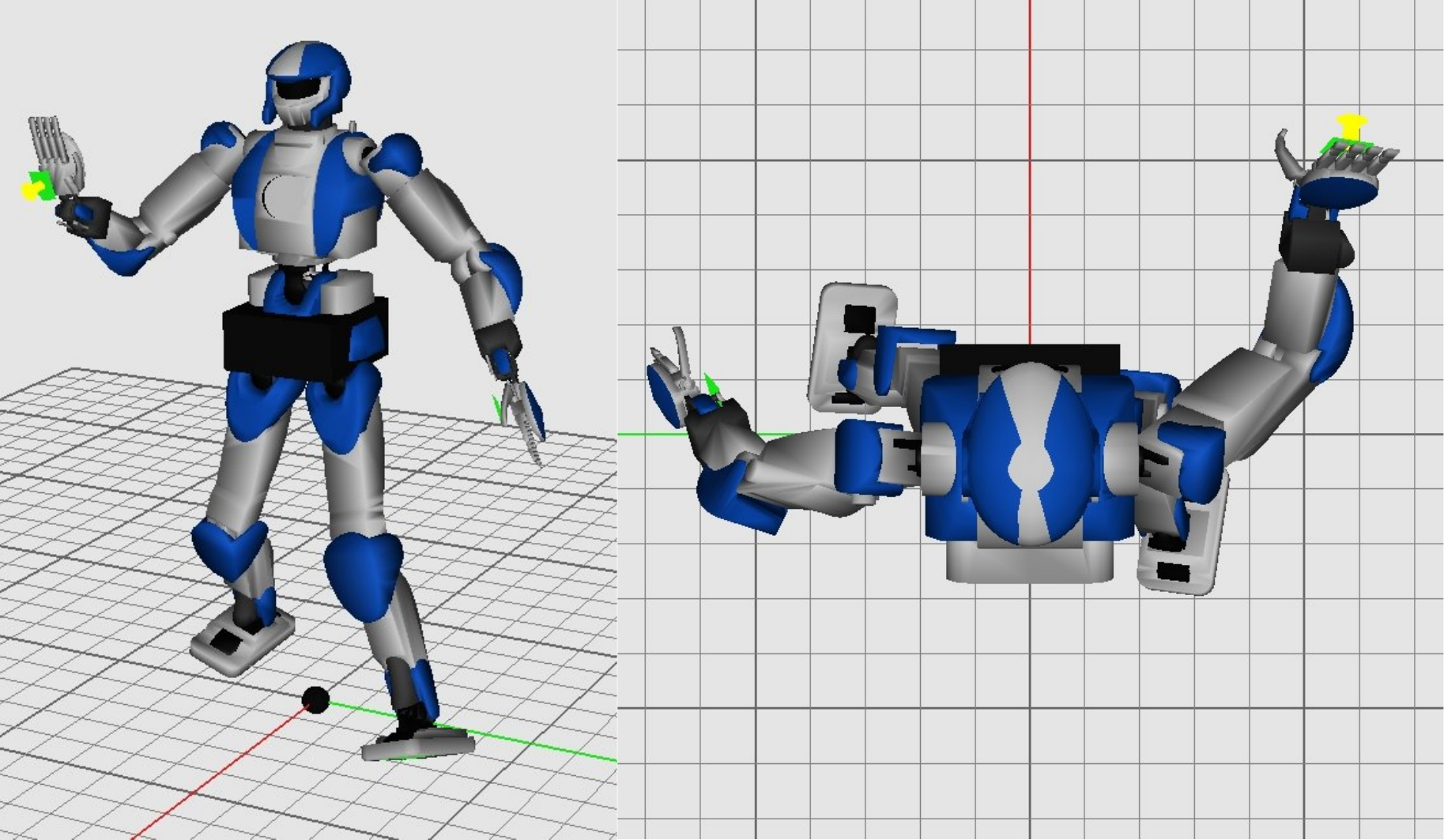}
  \caption{
    The HRP-4 robot balance with another two co-planar contacts.
  }
  \label{fig:double-contacts}
\end{figure}

\begin{figure}[htbp!]
  \centering
  \includegraphics[width=0.8\columnwidth]{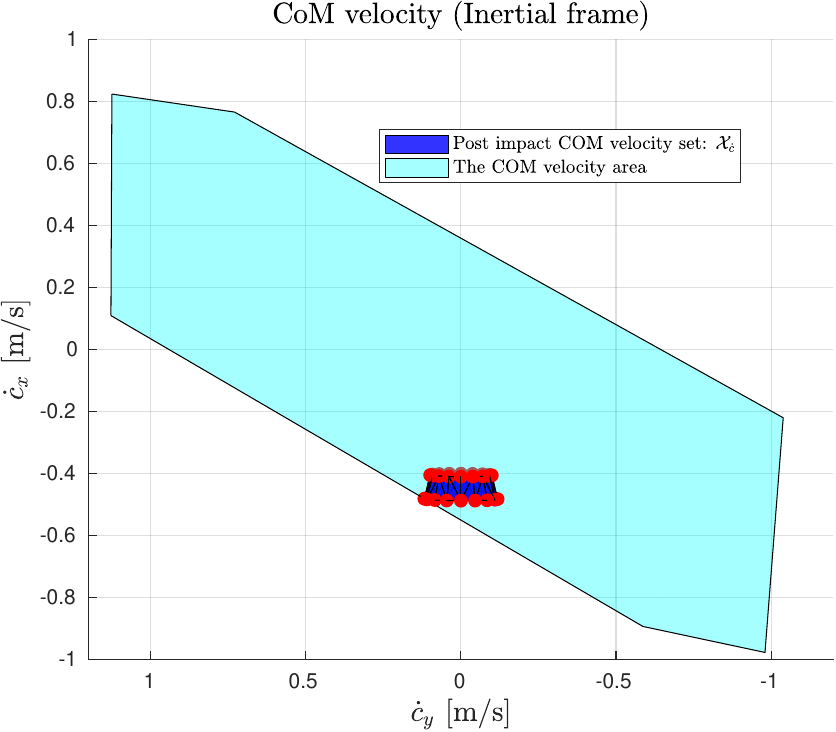}
  \caption{
    When the contact velocity in Fig.~\ref{fig:double-contacts} reached $\jacobian\optimal{\jvelocities}=0.529$\unitVelTS, a vertex of the set $\set{\comd}$ reaches the boundaries of $\comdArea$.
  }
  \label{fig:comd-double-contacts}
\end{figure}

\paragraph{Balance with two non-coplanar contacts}
The two non-coplanar foot contacts  with a friction coefficient $\coefF = 0.7$ in \figRef{fig:double-contacts}  lead to a significantly larger COM velocity area in \figRef{fig:comd-double-contacts} compared to \figRef{fig:comd-plot}.

\begin{figure}[htbp!]
  \centering
  \includegraphics[width=0.5\columnwidth]{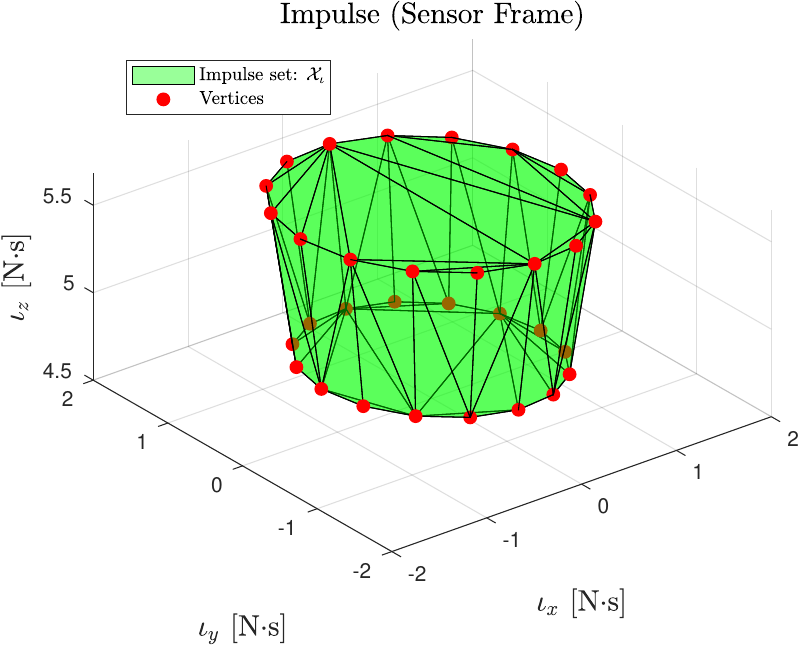}
  \includegraphics[width=0.38\columnwidth]{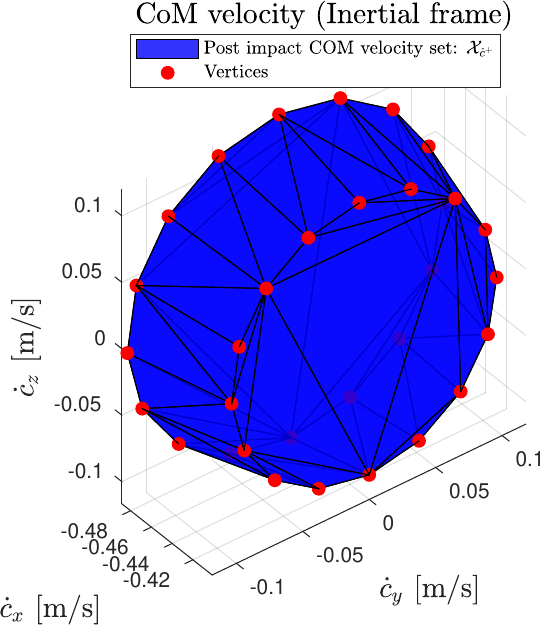}
  \caption{
    For the HRP-4 robot given in Fig.~\ref{fig:example_one}, we visualize vertices (red dots) of impulse set $\set{\impulse}$ and the COM velocity set $\set{\comd}$ when the contact velocity $\jacobian\optimal{\jvelocities}=0.529$\unitVelTS. The friction coefficient at the impact surface is $0.2$, and the restitution coefficients are limited to $\coefR \in [0, 0.2]$.
}
\label{fig:impulse_set_double_contacts}
\end{figure}

To evaluate the maximum contact velocity, we kept the right palm as the impact end-effector with the same friction coefficient $\coefF = 0.2$ and restitution coefficient $\coefR \in [0, 0.2]$.
Solving the QP \eqref{qp:velocity} generated a higher contact velocity of $0.529$ \unitVelTS, which corresponds to the \emph{overlapping condition} illustrated in \figRef{fig:comd-double-contacts} and the sets $\set{\impulse}$ and $\picomdSet$ in \figRef{fig:impulse_set_double_contacts}.


\begin{figure}[htbp!]
  \centering
  \includegraphics[height=5cm]{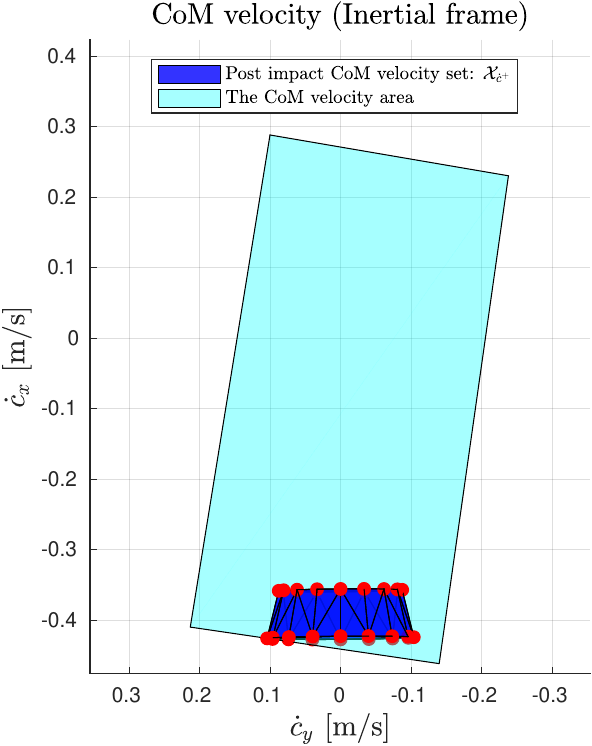}
  \includegraphics[height=5cm]{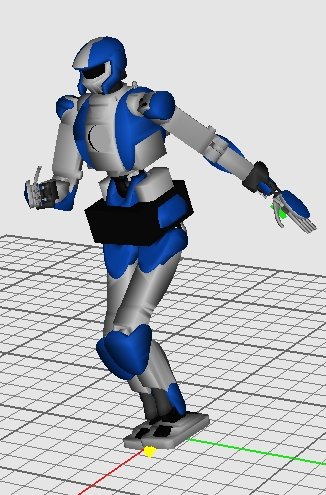}
  \caption{
    When the left toe kick at $\jacobian\optimal{\jvelocities}=0.569$\unitVelTS, a vertex of the set $\set{\comd}$ hits the boundaries of $\comdArea$.
  }
  \label{fig:comd-kicking}
  \vspace{-3mm}
\end{figure}

\begin{figure}[htbp!]
  \centering
  \includegraphics[width=0.5\columnwidth]{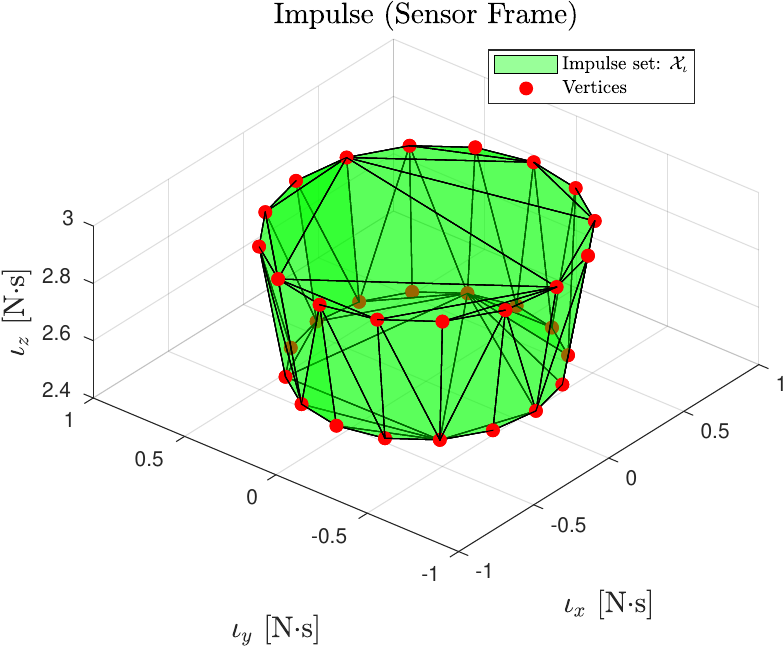}
  \includegraphics[width=0.4\columnwidth]{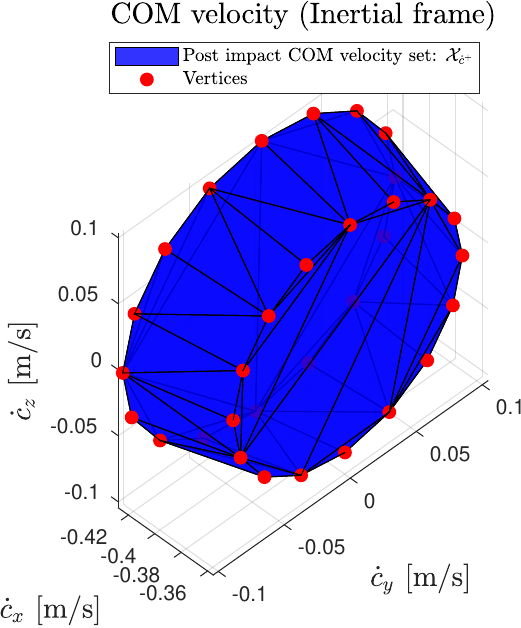}
  \caption{
    When the HRP-4 robot shown in Fig.~\ref{fig:comd-kicking} kicks at$\jacobian\optimal{\jvelocities}=0.569$\unitVelTS, we visualize vertices (red dots) of impulse set $\set{\impulse}$ and the COM velocity set $\set{\comd}$. The friction coefficient at the impact surface is $0.24$, and the restitution coefficients are limited to $\coefR \in [0, 0.2]$.
}
\label{fig:impulse_set_kicking}
  \vspace{-3mm}
\end{figure}

\paragraph{Kicking}
Employing the right foot contact with friction coefficient $\coefF = 0.7$, the HRP-4 robot kicks with the left foot's toe, see \figRef{fig:comd-kicking}.
We set the friction coefficient at the impact point as $\coefF = 0.24$ and restitution coefficient  $\coefR \in [0, 0.2]$.
The impact-aware QP \eqref{qp:velocity} found the \emph{maximum contact velocity} at $0.569$\unitVelTS,
which resulted in the sets $\set{\impulse}$ and $\picomdSet$ depicted in \figRef{fig:impulse_set_kicking}, and satisfied the \emph{overlapping condition} in \figRef{fig:comd-kicking}.

\section{Conclusion} 
\label{sec:conclusion}

% (1) Summarize the main findings

In order to deploy humanoid robots in a contact-rich environment, the balance criteria has to handle uneven ground and on-purpose impact tasks.
Our observations of the full-size humanoid robot HRP-4 indicated that the state-of-the-art ZMP-based balance criteria is overly conservative and ill-defined in impulse dynamics. 

To address this problem, we propose a CoM-velocity-based balance criteria for humanoid robots on non-coplanar contacts, and an optimization-based formulation that can solve for the maximum contact velocity while fulfilling the criteria.
Our assumptions include (1) the contacts are rigid, (2) the robot is kinematic-controlled, (3) the robot can timely detect the collision such that the impact will be instantaneous without exerting additional momentum exchange, and (4) the sustained contacts do not break during the instantaneous impact event.

We validated our balance criteria through a push-recovery experiment on the HRP-4 robot,
which sustained one foot on the ground and the other on a 30-degree ramp. 
To the best of our knowledge, this is the first successful push-recovery experiment for kinematic-controlled humanoid robot on non-coplanar contacts.
Additionally, we  evaluated the maximum contact velocities for other stances through simulations.

% (2) Re-state the research problem
% (3) Significance and implications, .. .. ..
Our approach has significant potential in enabling humanoid robots to perform various impact tasks with greater standing stability in contact-rich environments.
By determining the maximum foot contact velocities for different stances, we can plan and execute dynamic reference trajectories without breaking balance through impacts.

% (4) Limitations and future research
We acknowledge that the experimental validation of our approach is limited by the fragility of the HRP-4 robot, which has a payload of only $500$~grams. 
In future work, we will validate our approach on more robust platforms.
% Additionally, we plan to incorporate 3D template models such as the spring-loaded model~\cite{wensing2013iros} or the variational height model~\cite{caron2019tro}.
Furthermore, we aim to investigate damping post-impact oscillations through compliant control modes.

\appendix
\subsection{Resultant torque}
The wrench  $\inertialWrench = \vectorTwoRow{\inertialForce}{\inertialTorque}$ 
at the origin $\point{\inertialFrame}$ of the inertial frame will lead to torque at another point  $\point{x}$ as:
\quickEq{eq:resultant-torque}{
  \torque_{\point{x}} =  \resultantTorqueTwo{\inertialForce}{\inertialTorque}{\point{\inertialFrame}}{\point{x}}
  = 
  \resultantTorque{\inertialForce}{\inertialTorque}{\point{\inertialFrame}}{\point{x}}.
}
\subsection{Derivation of ZMP}
\label{app:cwc}
According to ZMP definition, the resultant torque at ZMP $\torque_{\zmp}$ is parallel to the ground surface normal $\nz$. Hence, we can derive the ZMP by substituting   $\torque_{\zmp}$ according to \eqref{eq:resultant-torque} into $\nz \times \torque_{\zmp} = 0$:  
\begin{equation}
\label{eq:derivation_zmp}
\begin{aligned}
\nz \times \torque_{\zmp} &= \nz \times (\resultantTorque{\inertialForce}{\inertialTorque}{\point{\inertialFrame}}{\zmp}) =0\\
\point{\inertialFrame} = 0 \Rightarrow &= \nz \times \inertialTorque - \nz \times (\zmp \times \inertialForce) = 0\\
& = \nz \times \inertialTorque - (\innerP{\nz}{\inertialForce})\zmp  + (\innerP{\nz}{\zmp})\inertialForce= 0\\
\Rightarrow \zmp &\defeq \frac{\nz \times \inertialTorque}{\innerP{\nz}{\inertialForce}}  +  \frac{ (\innerP{\nz}{\zmp}) \inertialForce}{\innerP{\nz}{\inertialForce}} \\ 
&= \frac{\nz \times \inertialTorque}{\innerP{\nz}{\inertialForce}}  +  \height{\zmp} \frac{\inertialForce}{\innerP{\nz}{\inertialForce}}.
\end{aligned}
\end{equation}

% Resultant wrench at
\subsection{Contact wrench}

The Newton-Euler equations of the robot writes:
\begin{equation}
\label{eq:newton-euler}
\begin{bmatrix}
\dot{\linearMomentum} \\
\dot{\angularMomentum}
\end{bmatrix}  = 
\begin{bmatrix}
\mass \ddot{\com} \\
\dot{\angularMomentum}
\end{bmatrix} 
 = 
\begin{bmatrix}
\mass \gForce \\
0
\end{bmatrix} 
  + \agg{i}{\numContacts}{ \begin{bmatrix}
      \force_{i} \\
      \pointer{\com}{\contactPoint_i} \times \force_{i}
\end{bmatrix}},
\end{equation}
where 
$m$ denotes the total mass of the robot. The vector $\pointer{\com}{\contactPoint_i} = \pointerDef{\com}{\contactPoint_i}$ connects a contact point $\contactPoint_i$ to the CoM.
Thus, $\force_{i}$ induces the torque
$\pointer{\com}{\contactPoint_i} \times \force_{i}$ at the CoM.

According to definition \eqref{eq:sum_wrench}, 
the $\numContacts$ sustained contact forces result in the  wrench $\inertialWrench$
at the inertial frame's origin $\point{\inertialFrame}$:
%% We can define the \emph{contact wrench} in the inertial frame as the sum of the resultant wrench : 
 \begin{equation}
   \label{eq:contact_wrench}
   \inertialWrench = \wrenchC{\inertialForce}{\inertialTorque} = 
\agg{i}{\numContacts}{ \begin{bmatrix}
    \force_{i} \\
    \pointer{\point{\inertialFrame}}{\contactPoint_i} \times \force_{i}
\end{bmatrix}}
=
\agg{i}{\numContacts}{ \begin{bmatrix}
    \force_{i} \\
    \contactPoint_i \times \force_{i}
\end{bmatrix}}
.
\end{equation}

 According to the equations of motion \eqref{eq:newton-euler}, the resultant force is equal to:
 $$
 \inertialForce = \linearMomentumD - \mass\gForce,
 $$
and we can expand $\angularMomentumD$ as:
%%  By substituting $\angularMomentumD = \agg{i}{\numContacts}{\pointer{\com}{\contactPoint_i} \times \force_{i}}$ from \eqref{eq:newton-euler}, we can  expand $\inertialTorque$ as:
 \quickEq{eq:angularmomentum-expand}{
 \begin{aligned}
\angularMomentumD  & = \agg{i}{\numContacts}{\pointer{\com}{\contactPoint_i} \times \force_{i}}
 = \agg{i}{\numContacts}{
   (
   \pointerDef{
     \com}
              {\contactPoint_i}
              )
              \times \force_{i}} \\
 &=
 \underbrace{\agg{i}{\numContacts}{
   \contactPoint_i \times \force_{i}}}_{\inertialTorque}
 -   \com \times
 \underbrace{\agg{i}{\numContacts}{\force_{i}}}_{\inertialForce}
 \\
 &=  \inertialTorque - \com \times\inertialForce.
 \end{aligned}
}

\subsection{Derivation of the whole-body dynamics \eqref{eq:three_relation}}
\label{app:com-zmp}
By substituting $\inertialForce = \linearMomentumD - \mass\gForce$ and $\inertialTorque = \angularMomentumD + \com \times \inertialForce$ according to \eqref{eq:angularmomentum-expand}, 
we can re-write the ZMP definition \eqref{eq:def_zmp}: 
\begin{equation*}
\begin{aligned}
(\innerP{\nz}{\inertialForce}) \zmp &=
  \nz \times \inertialTorque  +   (\innerP{\nz}{\zmp}) \inertialForce \\
  &= \nz \times 
(\angularMomentumD + \com \times \inertialForce)
  +   (\innerP{\nz}{\zmp}) \inertialForce \\ 
  &= \cross{\nz}{\dot{\angularMomentum}} + \cross{\nz}{(\com \times \inertialForce)}
  +   (\innerP{\nz}{\zmp}) \inertialForce  \\
  &= \cross{\nz}{\dot{\angularMomentum}} + \vectripleexpan{\nz}{\com}{\inertialForce} + (\innerP{\nz}{\zmp}) \inertialForce \\
(\innerP{\nz}{\inertialForce}) (\zmp  - \com )&=\cross{\nz}{\dot{\angularMomentum}} + 
(\innerP{\nz}{(\zmp - \com)})\inertialForce \\ 
(\innerP{\nz}{\inertialForce}) \pointer{\com}{\zmp}&=\cross{\nz}{\dot{\angularMomentum}} + 
(\innerP{\nz}{\pointer{\com}{\zmp}})\inertialForce \\ 
\innerP{\nz}{(\dot{\linearMomentum} - \mass\gForce)} \pointer{\com}{\zmp}&=
\cross{\nz}{\dot{\angularMomentum}} + (\innerP{\nz}{\pointer{\com}{\zmp}})(\dot{\linearMomentum} - \mass\gForce) \\ 
\innerP{\nz}{(\mass \ddot{\com} - \mass\gForce)} \pointer{\com}{\zmp}&=
\cross{\nz}{\dot{\angularMomentum}} + (\innerP{\nz}{\pointer{\com}{\zmp}})(\mass \ddot{\com} - \mass \gForce) \\ 
\Rightarrow \mass (\innerP{\nz}{\pointer{\com}{\zmp}}) \ddot{\com} 
&=  \mass (\innerP{\nz}{\pointer{\com}{\zmp}})\gForce - \cross{\nz}{\dot{\angularMomentum}}\\
& + \innerP{\nz}{(\mass \ddot{\com} - \mass\gForce)} \pointer{\com}{\zmp} \\
\Rightarrow 
\ddot{\com} &= \gForce
- \frac{\cross{\nz}{\dot{\angularMomentum}}}{\mass (\innerP{\nz}{\pointer{\com}{\zmp}})} 
+ \frac{\innerP{\nz}{(\mass \ddot{\com} - \mass\gForce)} \pointer{\com}{\zmp}}{\mass (\innerP{\nz}{\pointer{\com}{\zmp}})} \\
~\Rightarrow 
\ddot{\com} &= \underbrace{\gForce + \frac{\cross{\dot{\angularMomentum}}{\nz}}{\mass (\height{\com} - \height{\zmp})} 
+ \frac{\innerP{\nz}{( \ddot{\com} - \gForce)} \pointer{\com}{\zmp} }{ 
  \height{\com} - \height{\zmp}}}_{
  \innerP{\nz}{\pointer{\com}{\zmp}} = \height{\com} - \height{\zmp}
}.
\end{aligned}
\end{equation*}
\subsection{Derivation of the LIPM dynamics \eqref{eq:lipm-model-3d}}
\label{app:lipm-model-3d}

By substituting the two LIPM assumptions \eqref{eq:lipm_ass_1} and \eqref{eq:lipm_ass_2} into the whole-body dynamics \eqref{eq:three_relation}, we can simplify it to the LIPM dynamics \eqref{eq:lipm-model-3d}:
$$
\begin{aligned}
\comdd &= 
\gForce - \frac{\innerP{\nz}{\gForce} \pointer{\com}{\zmp}}{\height{\com} - \height{\zmp}} 
= \gForce  - \frac{g \pointer{\com}{\zmp}}{\height{\com} - \height{\zmp}} 
= \gForce  - \frac{g }{\height{\com} - \height{\zmp}} (\zmp - \com)   \\
&= \gForce  + \frac{g }{\height{\com} - \height{\zmp}} (\com - \zmp).
\end{aligned}
$$

\subsection{The contact wrench cone}
\label{app:cwc}
Caron et al.~\cite{caron2015icra} established that for a given friction coefficient  $\mu$, a range of limited rotational torque $\torque^z \in [\bs{\tau}^z_{\text{min}}, \bs{\tau}^z_{\text{max}}]$ and the geometric size $[X, Y]$\unitPos, the $i$-th planar contact can maintain a stationary state if its contact wrench $ \wrenchFrame{i}{i} = \vectorTwoRow{\torque}{\force}$ (represented in its local coordinate frame) satisfies the following condition:
\begin{align*}
 |\bs{f}^x | & \leq \mu  \bs{f}^z,  \\
 |\bs{f}^y | & \leq \mu  \bs{f}^z,  \\ 
 |\bs{f}^z | & > 0, \\
 |\bs{\tau}^x | & \leq Y  \bs{f}^z, \\
 |\bs{\tau}^y | & \leq X  \bs{f}^z, \\
\bs{\tau}^z_{\text{min}} & \leq  \bs{\tau}^z \leq \bs{\tau}^z_{\text{max}}. 
\end{align*}
The half-space representation of the above inequalities write:
\begin{equation}
\label{eq:cwc_local}
\cwcFrame{i}{i} \wrenchFrame{i}{i} \leq 0,
\end{equation}
where $\cwcFrame{i}{i}$ is given by: 
$$
\begin{bmatrix}
  0,   &0,   &0,  &-1,   &0,          & -\mu \\
  0,   &0,   &0,  &1,   &0,           & -\mu \\
  0,   &0,   &0,   &0,  &-1,          & -\mu \\
  0,   &0,   &0,   &0,  &+1,          & -\mu \\
  -1,  &0,   &0,   &0,  &0,           & -Y \\ 
  +1,  &0,   &0,   &0,  &0,           & -Y \\
  0,  &-1,   &0,   &0,   &0,          & -X  \\
  0,  &+1,   &0,   &0,   &0,          & -X  \\ 
   +\mu, &+\mu,  &-1,  &-Y,  &-X, &-(X + Y) \mu, \\
  +\mu, &-\mu,  &-1,  &-Y,  &+X, &-(X + Y) \mu, \\
  -\mu, &+\mu,  &-1,  &+Y,  &-X, &-(X + Y) \mu, \\
  -\mu, &-\mu,  &-1,  &+Y,  &+X, &-(X + Y) \mu, \\
  +\mu, &+\mu,  &+1,  &+Y,  &+X, &-(X + Y) \mu, \\
  +\mu, &-\mu,  &+1,  &+Y,  &-X, &-(X + Y) \mu, \\
  -\mu, &+\mu,  &+1,  &-Y,  &+X, &-(X + Y) \mu, \\
  -\mu, &-\mu,  &+1,  &-Y,  &-X, &-(X + Y) \mu 
\end{bmatrix}.
$$
As we need to formulate \eqref{eq:cwc_local} with respect to  wrench $\wrench_i$, whose orientation aligns with inertial frame $\cframe{\inertialFrame}$, we modify the constraint \eqref{eq:cwc_local}  as: 
$$
\underbrace{\cwcFrame{i}{i}
  \matrixTwo{\rotation{\inertialFrame}{i}}{0}{0}{\rotation{\inertialFrame}{i}}
  }_{
  \cwc_i
}\wrench_i \leq 0.
$$

\subsection{Equations of motion}
\label{app:eom}
A floating-base robot has $\ajDim$
actuated joints $\jointPosition \in \RRv{\ajDim}$ and
$6$ under-actuated floating-base joints in $\RRv{6}$. Thus, the total 
degrees of freedom (DOF) is $\jsDim$.
Assuming $\numContacts$  sustained contacts, the equations of motion write:
\begin{equation}
  \label{eq:eom}
  \inertiaMatrix(\jangles)\jaccelerations  + \gravityandcoriolis(\jangles, \jvelocities) =  \actuationMatrix \jtorques +
  \transpose{\jacobian}\wrench, 
\end{equation}
where
$\inertiaMatrix(\jangles)\in \RRm{\jsDim}{\jsDim}$ is the joint-space inertia matrix,
$\gravityandcoriolis(\jangles, \jvelocities) \in \mathbb{R}^{\jsDim}$ gathers both Coriolis and gravitation vectors.
We drop the dependency on $\jangles$ and $\jvelocities$ in the rest of the paper for simplicity.
$\actuationMatrix \in \RRm{\jsDim}{\ajDim}$ selects the actuated joints from $\jangles$; $\jtorques \in \mathbb{R}^{\ajDim}$ denotes the joint torques.

% Define the enw
Further, $\jacobian \in \RRm{6\numContacts}{\jsDim}$ and $\wrench \in \RRm{6\numContacts}{1}$ vertically stack $\numContacts$
sustained contacts' Jacobians  and wrenches, respectively\footnote{We are free to choose any coordinate frame to represent both $\jacobian$  and $\wrench$. }:
$$
\jacobian = \begin{bmatrix}
\jacobian_1 \\ 
\vdots\\ 
\jacobian_{\numContacts}
\end{bmatrix},
\quad 
\wrench = \begin{bmatrix}
\wrench_1 \\ 
\vdots\\ 
\wrench_{\numContacts}
\end{bmatrix}.
$$
Each pair of the Jacobian and the wrench $\jacobian_i, \wrench_i$ for $i = 1,\ldots,\numContacts$ aligns with the inertial frame\footnote{Suppose $\wrenchFrame{i}{i}$ denotes
the $i$th contact's wrench in the local frame $\cframe{i}$,
its counterpart in the inertial frame is
$\wrench_i = \matrixTwo{\rotation{\inertialFrame}{i}}{0}{0}{\rotation{\inertialFrame}{i}} \wrenchFrame{i}{i}$
%% $$
%% \geometricFT{i}{\inertialFrame} = \geometricFTDef{i}{\inertialFrame}
%% $$
}. 

\subsection{Derivation of  the LIPM equality constraint \eqref{eq:angular_momentum_assumption_constraint}}
\label{app:com-moment}
Given the wrench $\inertialWrench$, we can re-write the torque about the CoM $\torque_{\comFrame}$ according to \eqref{eq:resultant-torque}:
\begin{equation}
\label{eq:torque_com}
\torque_{\comFrame} = \resultantTorque{\inertialForce}{\inertialTorque}{\point{\inertialFrame}}{\com} = \inertialTorque - \com \times \force.  
\end{equation}
By substituting \eqref{eq:torque_com}, we can re-write the LIPM assumption $\innerP{\nz}{\torque_{\comFrame}} = 0$ \eqref{eq:lipm_ass_1} as: 
\begin{equation}
\label{eq:expansion_assumption1}
\begin{aligned}
\innerP{\nz}{\inertialTorque} - \innerP{\nz}{(\com \times \inertialForce)} &= 0,  \\
\innerP{\nz}{\inertialTorque} - \innerP{\inertialForce}{( \nz \times \com )} &= 0,  \\
\innerP{\nz}{\inertialTorque} - \innerP{( \nz \times \com )}{\inertialForce} &= 0.  
\end{aligned}
\end{equation}
where we \emph{circular-shiftted} the scalar triple product and \emph{swapped} the operator. Similarly, we can re-write the other LIPM assumption $\cross{\nz}{\torque_{\comFrame}} = 0$ \eqref{eq:lipm_ass_2} by substituting \eqref{eq:torque_com} as: 
\begin{equation}
\label{eq:expansion_assumption3}
\begin{aligned}
\cross{\nz}{\inertialTorque} - \cross{\nz}{(\com \times \inertialForce)} &= 0, \\
\cross{\nz}{\inertialTorque} - (\innerP{\nz}{\inertialForce})\com + (\innerP{\nz}{\com})\inertialForce &= 0,  \\
\cross{\nz}{\inertialTorque} - \mass g\com + \height{\com}\inertialForce &= 0,  \\
\cross{\nz}{\inertialTorque}  + \height{\com}\inertialForce &= \mass g\com.
\end{aligned}
\end{equation}
Collecting \eqref{eq:expansion_assumption1} and \eqref{eq:expansion_assumption3} in a matrix form, we obtain the equality constraints on the wrench $\inertialWrench = \wrenchR{\inertialForce}{\inertialTorque}$: 
$$
\frac{1}{\mass g} 
\begin{bmatrix} 
  \height{\com} \cdot \idMatrix{3}  &\hat{\nz}\\
-(\cross{\nz}{\com})^\top & \nz^\top 
\end{bmatrix} 
\wrenchC{\inertialForce}{\inertialTorque} =
\begin{bmatrix}
\com \\
0
\end{bmatrix}.
$$

\subsection{Derivation of  ZMP dependence on CoM and external forces \eqref{eq:mc_zmp_projection}}
\label{app:zmp-com-force}
The wrench $\inertialWrench = \wrenchR{\inertialForce}{\inertialTorque}$  will result in the torque at ZMP as:
% we have the torque about $\zmp$: 
$$
\torque_{\zmp} = \resultantTorque{\inertialForce}{\inertialTorque}{\point{\inertialFrame}}{\zmp}.
$$
Substituting $\torque_{\zmp}$  into the condition $\nz \times \torque_{\zmp} = 0$ and  expand the vector triple product, we find that ZMP $\zmp$ fulfills:
\begin{equation}
\label{eq:zmp_def}
\begin{aligned}
\nz \times \torque_{\zmp} &= \nz \times (\resultantTorque{\inertialForce}{\inertialTorque}{\point{\inertialFrame}}{\zmp}) =\zeroVector\\
&= \nz \times \inertialTorque - \nz \times (\zmp \times \inertialForce) = \zeroVector% \\
% & = \bs{n} \times \torque_O - (\inner{\bs{n}}{\force})\zmp  + (\inner{\bs{n}}{\zmp})\force= \zeroVector\\
\end{aligned}.
\end{equation}
Equating \eqref{eq:zmp_def} to
$
\cross{\nz}{\torque_{\comFrame}} %% - \cross{\nz}{{\angularMomentumD}_{\com}}
= \bs{0}
$, which is obtained by  re-arranging \eqref{eq:lipm_ass_3}, we can write ZMP as a function of the CoM and $\inertialForce$:
$$
  \zmp = \com + \frac{\height{\zmp} - \height{\com}}{\mass g}\inertialForce. % + \frac{1}{mg}\cross{\bs{n}}{{\angularMomentumD}_{\com}}.
$$

\section*{Acknowledgment}
%This work was partially supported by the EU H2020 research grant GA 871899, I.AM. project.
We thank Pierre Gergondet for his continuous support in setting up the \href{https://github.com/jrl-umi3218/mc_rtc}{\tt mc\_rtc} controller, St{\'e}phane Caron for the critical feedback on multi-contact ZMP area, Saeid Samadi for the HRP-4 experiment, and Julien Roux for the C++ implementation of~\cite{bretl2008tro,audren2018tro} \href{https://github.com/JulienRo7/stabiliplus}{\tt stabiliplus}.
%% , and Ahmed Zermane for the proof reading.

\bibliographystyle{IEEEtran}
\bibliography{ref}

\end{document}